%% file: iclr2026_arxiv.tex
\documentclass{article} 
\usepackage{iclr2026_conference}
\usepackage{times}

\input{math_commands.tex}

\usepackage{microtype}
\usepackage[pdftex]{graphicx}
\usepackage{subfiles}
\usepackage{hyperref}
\usepackage{url}
\usepackage{booktabs}
\usepackage{tcolorbox}
\usepackage{float}
\usepackage{lineno}
\usepackage{amsmath}
\usepackage{multirow}
\usepackage{subcaption}
\usepackage{enumitem}
\usepackage{caption}
\usepackage{arydshln}
\definecolor{darkblue}{rgb}{0, 0, 0.5}
\hypersetup{colorlinks=true, citecolor=darkblue, linkcolor=darkblue, urlcolor=darkblue}

\usepackage{xcolor}
\usepackage{tikz}
\usetikzlibrary{calc}
\definecolor{lightred}{rgb}{0.78, 0, 0}
\definecolor{lightgreen}{rgb}{0, 0.78, 0}
\definecolor{lightblue}{rgb}{0, 0, 0.78}
\definecolor{copy}{rgb}{0, 0.5, 0}
\definecolor{done}{rgb}{1, 0.28, 0.31}
\definecolor{todo}{rgb}{0, 0, 0.5}
\definecolor{grey}{rgb}{0.7, 0.7, 0.7}

\title{JudgeLRM: Large Reasoning Models as a Judge}

\def\huggingface{\raisebox{-1.5pt}{\includegraphics[height=1.05em]{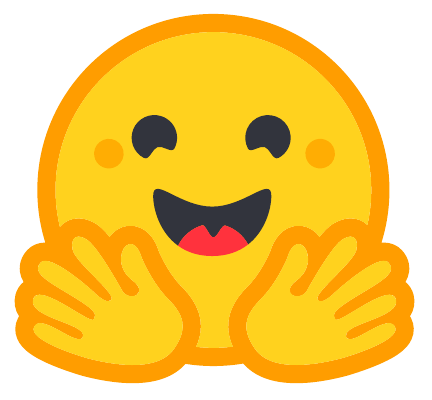}}}
\def\github{\raisebox{-1.5pt}{\includegraphics[height=1.05em]{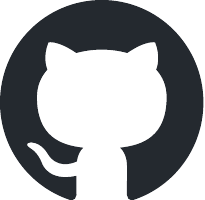}}}
\def\spaces{\raisebox{-1.5pt}{\includegraphics[height=1.05em]{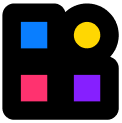}}}

\iclrfinalcopy

\author{Nuo Chen \ \ \ \ \
 Zhiyuan Hu \quad Qingyun Zou \quad \textbf{Jiaying Wu} \\
   \textbf{Qian Wang} \quad
   \textbf{Bryan Hooi} \quad
  \textbf{Bingsheng He}
\\
National University of Singapore\\
 \href{mailto:nuochen@comp.nus.edu.sg}{nuochen@comp.nus.edu.sg}\\
 \vspace{-9mm}
}
\begin{document}

\maketitle

\begin{abstract}
\vspace{-3mm}
Large Language Models (LLMs) are increasingly adopted as evaluators, offering a scalable alternative to human annotation. However, existing supervised fine-tuning (SFT) approaches often fall short in domains that demand complex reasoning. Judgment is inherently reasoning-intensive: beyond surface-level scoring, it requires verifying evidence, identifying errors, and justifying decisions. Through the analysis of evaluation tasks, we find a negative correlation between SFT performance gains and the proportion of reasoning-demanding samples, revealing the limits of SFT in such scenarios. To address this, we introduce \textbf{JudgeLRM}, a family of judgment-oriented LLMs, trained using reinforcement learning (RL) with judge-wise, outcome-driven rewards to activate reasoning capabilities. JudgeLRM consistently outperform SFT-tuned baselines in the same size, as well as other RL and SFT variants, and even surpass state-of-the-art reasoning models: notably, JudgeLRM-3B/4B exceeds GPT-4, while JudgeLRM-7B/8B/14B outperforms DeepSeek-R1 by over 2\% in F1 score, with particularly strong gains on reasoning-heavy tasks. Our findings underscore the value of RL in unlocking reasoning-aligned LLM judges.
\end{abstract}

\newcommand{\plus}[1]{\textcolor{green!60!black}{\scriptsize\,+#1}}
\newcommand{\minus}[1]{\textcolor{red!70!black}{\scriptsize\,-#1}}
\vspace{-1mm}
\begin{table}[h]
\begin{center}
\resizebox{\textwidth}{!}{
\begin{tabular}{lcccccccc}
\toprule
\multicolumn{1}{l}{Dataset} & \multicolumn{4}{c}{\textit{JudgeLM (GPT-4 as ground truth. In distribution.)}} & \multicolumn{4}{c}{\textit{PandaLM (Human as ground truth. Out of distribution.)}} \\
\cmidrule(r){2-5} \cmidrule(r){6-9}
\multirow{1}{*}{\textbf{Criteria}} & \multicolumn{1}{c}{\textbf{Agreement}} & \multicolumn{1}{c}{\textbf{Precision}} & \multicolumn{1}{c}{\textbf{Recall}} & \multicolumn{1}{c}{\textbf{F1}} & \multicolumn{1}{c}{\textbf{Agreement}} & \multicolumn{1}{c}{\textbf{Precision}} & \multicolumn{1}{c}{\textbf{Recall}} & \multicolumn{1}{c}{\textbf{F1}} \\
\hline \quad Qwen2.5-3B-Instruct  & 72.29 & 80.19 & 64.07 & 71.23 & 68.50 & 50.92 & 56.13 & 51.57 \\
\quad Qwen2.5-3B-Instruct-Judge-SFT
& 83.58\plus{\textbf{11.29}} & 75.25\minus{\textbf{4.94}} & 76.12\plus{\textbf{12.05}} & 75.05\plus{\textbf{3.82}}
& 70.57\plus{\textbf{2.07}} & 67.09\plus{\textbf{16.17}} & 73.36\plus{\textbf{17.23}} & 66.10\plus{\textbf{14.53}} \\
\quad JudgeLRM-3B
& 83.72\plus{\textbf{11.43}} 
& 86.31\plus{\textbf{6.12}} 
& 82.98\plus{\textbf{18.91}} 
& \textbf{84.61}\plus{\textbf{13.38}}
& 77.68\plus{\textbf{9.18}} 
& 74.26\plus{\textbf{23.34}} 
& 70.86\plus{\textbf{14.73}} 
& \textbf{72.12}\plus{\textbf{20.55}} \\
\hdashline
\quad Qwen2.5-7B-Instruct  & 76.85 & 78.71 & 77.85 & 78.28 & 63.96 & 61.95 & 67.61 & 59.81 \\
\quad Qwen2.5-7B-Instruct-Judge-SFT
& 82.00\plus{\textbf{5.15}} & 84.43\plus{\textbf{5.72}} & 81.74\plus{\textbf{3.89}} & 83.06\plus{\textbf{4.78}}
& 73.57\plus{\textbf{9.61}} & 67.31\plus{\textbf{5.36}} & 72.23\plus{\textbf{4.62}} & 67.98\plus{\textbf{8.17}} \\
\quad JudgeLRM-7B
& 83.74\plus{\textbf{6.89}} 
& 85.84\plus{\textbf{7.13}} 
& 83.65\plus{\textbf{5.80}} 
& \textbf{84.73}\plus{\textbf{6.45}}
& 78.28\plus{\textbf{14.32}} 
& 74.90\plus{\textbf{12.95}} 
& 75.74\plus{\textbf{8.13}} 
& \textbf{75.05}\plus{\textbf{15.24}} \\
\bottomrule
\end{tabular}
}
\end{center}
\captionsetup{type=table}
\vspace{-2mm}
\caption{Performance improvement from base models to SFT and our judge-wise RL methods on JudgeLM (in-distribution training data) and PandaLM (out-of-distribution for generalization).}
\vspace{-2mm}
\label{tab:judgeneedreasoning}
\end{table}
\vspace{-1mm}
\section{Introduction}
Recent advancements in LLMs have sparked significant interest in their use as evaluative judges \citep{gu2025surveyllmasajudge,li2024llmsasjudgescomprehensivesurveyllmbased}, offering a reliable and scalable alternative to costly human annotation. Previous approaches, such as JudgeLM~\citep{zhu2025judgelm} and PandaLM~\citep{wang_pandalm_2024}, have enabled LLMs to perform judgment tasks based on the large scale  SFT.

\begin{figure}
    \centering
    \includegraphics[width=0.6\linewidth]{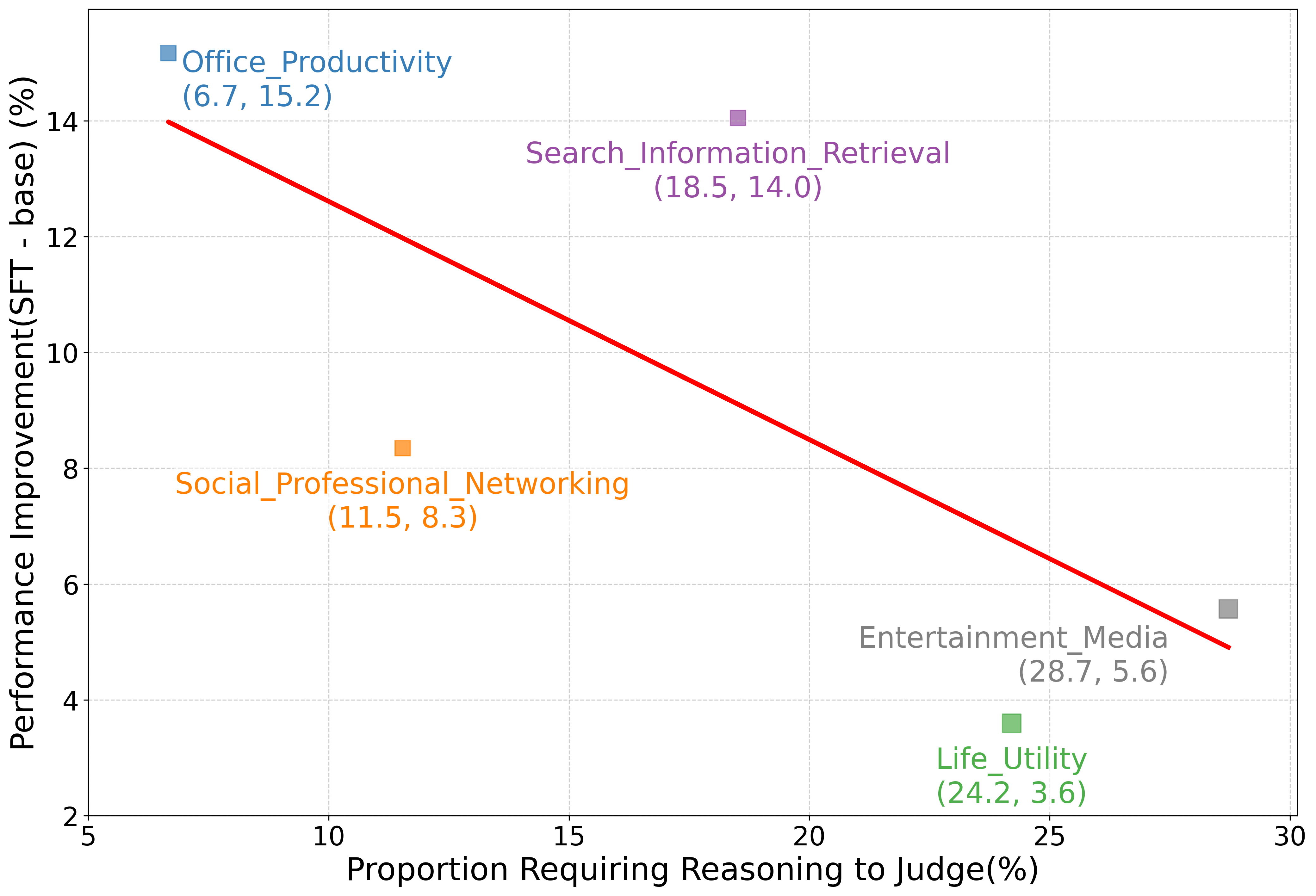}
    \caption{Judgment performance improvement vs. reasoning requirement across domainY-axis shows F1 score improvement (SFT – base) based on Qwen2.5-7B-instruct; X-axis shows the proportion of tasks requiring reasoning. Each point represents a domain. A negative linear trend (y = -0.41x + 16.72, R² = 0.53) suggests that domains with more reasoning-heavy tasks benefit less from SFT alone. Sample counts across domains: 105 (Office\_Productivity), 108 (Search\_Information\_Retrieval), 195 (Entertainment\_Media), and estimated 108 (Social\_Professional\_Networking), 190 (Life\_Utility).}
    \label{fig:judge_sftvsbase}
    \vspace{-4mm}
\end{figure}

However, SFT faces inherent limitations in generalization and reasoning depth. \textbf{Do LLM judges truly require strong reasoning capabilities?} To explore this, we revisit prior work such as PandaLM \citep{wang_pandalm_2024}, which shows that smaller models can be adapted for evaluation via post-training techniques. However, these methods still struggle with complex reasoning demands and computational inefficiency. We investigate whether improvements in judgment performance through SFT correlate with the proportion of questions that require reasoning to evaluate. Using the five question source categories defined in PandaLM (see Section \ref{sec:pandalm_category}), we compute the proportion of samples in each category that require reasoning (detailed in Table \ref{tab:judge_reasoning_analysis}). 
As shown in Figure \ref{fig:judge_sftvsbase} and Table \ref{tab:judgeneedreasoning}, we observe an inverse relationship between improvements in F1 score on the judge task and the proportion of reasoning-required samples (elaborated in Appendix \ref{app:necessityreasoning}), indicating that SFT struggles more on categories demanding higher reasoning depth. Specifically, SFT gains drop as the proportion of reasoning-intensive items rises, implying that judges require flexible generalization skills rather than memorization abilities. This suggests that \textbf{effective LLM judges must possess strong reasoning abilities} to handle diverse and complex evaluation scenarios, where rote patterns from training memorization data fall short.

A fundamental challenge in deploying LLMs as judges lies in their dependence on intrinsic reasoning capabilities. While the Chain-of-Thought (CoT)~\citep{wei2022chain} framework via SFT equips models to process nuanced information, achieving accurate and contextually grounded judgments remains non-trivial. Studies have shown that advanced large reasoning models (LRMs), such as DeepSeek-R1 \citep{deepseekai2025deepseekr1incentivizingreasoningcapability}, demonstrate strong performance in evaluation tasks by leveraging structured reasoning paths. To bridge this gap, we propose \textbf{JudgeLRM}, a family of LLMs trained using RL with judge-wise, outcome-driven rewards to enhance evaluative reasoning. The task-specific reward function integrates both structural and content-based components. The structural reward ensures well-formatted reasoning and answer sections, while the content reward aligns model judgments with ground-truth preferences through relation, absolute, and confidence-based metrics. This design promotes both faithful reasoning and accurate, confident scoring.

JudgeLRM model series (licensed under MG0-2.0\footnote{\url{https://www.modelgo.li/}}) ranges from 3B to 14B parameters, trained using Group Relative Policy Optimization (GRPO) ~\citep{shao2024deepseekmathpushinglimitsmathematical}. Empirical results demonstrate that JudgeLRM not only surpasses proprietary models like GPT-4 and DeepSeek-R1 but also outperforms SFT and RL baselines of comparable sizes, with an average improvement of 8.14\% in F1 score over SFT counterparts. Notably, JudgeLRM shows consistent gains even in categories with a high proportion of reasoning-required tasks (see Table \ref{fig:judge_ourvssft}, Figure \ref{fig:judge_ourvssft}), further illustrating its ability to overcome SFT's reasoning limitations. These implications highlight that \textbf{judgment is inherently a reasoning-intensive task}, not merely a scoring exercise, and that RL-based approaches can effectively instill the flexible generalization needed for robust judge models. Through systematic analysis, we demonstrate that reasoning patterns, such as verification, sub-goal setting, double checking, error identification, and decision justification, are crucial to the success of judgment tasks.

\vspace{-2mm}
\section{Related Work}
\vspace{-1mm}
\subsection{LLM Reasoning and Planning}
Modern large language models (LLMs) demonstrate remarkable reasoning abilities through their intrinsic Chain-of-Thought (CoT) mechanisms \citep{wei2022chain}. This capability can be elicited through prompting techniques \citep{kojima2022large} or explicitly integrated into model architectures like Openai o1~\citep{openai2024o1}, Deepseek R1 \citep{deepseekai2025deepseekr1incentivizingreasoningcapability,shao2024deepseekmathpushinglimitsmathematical}. Such structural enhancements enable LLMs to perform multi-step reasoning across diverse domains, like in medical, Huatuo-o1 \citep{chen2024huatuogpto1}, Medical-R1 \citep{lai2025medr1} and in finance, Fin-o1 \citep{qian2025fino1} and Fin-R1 \citep{liu2025finr1largelanguagemodel}. Our work \textbf{first focuses on energizing reasoning abilities for judges}, subsequently revealing the inherently reasoning-intensive nature of the judging task and validates the necessity and effectiveness of strong reasoning skills for judges.

\subsection{LLM as a Judge}
Human evaluation of LLM outputs is time-consuming, resource-intensive, and often inconsistent due to annotator subjectivity \citep{gu2024survey}. To address these limitations, researchers have explored using LLMs themselves as evaluators, a paradigm often referred to as 'LLM-as-a-Judge' \citep{zheng2024judging}. This approach promises more scalable and potentially cost-effective evaluation. Furthermore, recognizing the potential benefits of specialized models, some studies have focused on training dedicated LLMs specifically for the task of judging LLM outputs, aiming for improved accuracy and alignment with human preferences \citep{zhu2025judgelm,wang_pandalm_2024}. Despite its promise, the LLM-as-a-Judge approach faces various biases inherent in the judge LLMs themselves, which can compromise the fairness and reliability of the evaluations \citep{gallegos2024biasfairnesslargelanguage, benyou2024bias,dubois2025lengthcontrolledalpacaevalsimpleway}. We \textbf{first introduce reinforcement learning on training small-scale LLMs as a judge}, demonstrating results comparable to Deepseek-R1 and a controllable range of bias. More recent studies have also explored RL for judge alignment \citep{whitehouse2025j1, xu2025j4rlearningjudgeequivalent}, reflecting a growing consensus on the potential of RL-based methods for this task.

\section{Judge-wise Outcome Reward and RL Training}
Inspired by the insufficiency of SFT training revealed in Fig. \ref{fig:judge_sftvsbase}, we introduce Large Reasoning Models as a judge (JudgeLRM), applying RL on judge tasks through judge-wise outcome reward.

\begin{figure*}[!h]
\centering
\begin{tcolorbox}[colback=blue!5!white, colframe=blue!70!black, title=System Prompt for RL Training, width=1.0\textwidth, sharp corners=south]
\linespread{0.9}\selectfont
\texttt{\textbf{<|im\_start|>system}}\\You are a helpful assistant. The assistant first performs a detailed, step-by-step reasoning process in its mind and then provides the user with the answer. The reasoning process and answer are enclosed within \texttt{<think>} \texttt{</think>} and \texttt{<answer>} \texttt{</answer>} tags, respectively, i.e., \texttt{<think>} detailed reasoning process here, explaining each step of your evaluation for both assistants \texttt{</think>}\texttt{<answer>} answer here \texttt{</answer>}. Now the user asks you to judge the performance of two AI assistants in response to the question. Score assistants 1-10 (higher=better). Criteria includes helpfulness, relevance, accuracy, and level of detail. Avoid order, length, style or other bias. After thinking, when you finally reach a conclusion, clearly provide your evaluation scores within \texttt{<answer>} \texttt{</answer>} tags, i.e., for example,\texttt{<answer>}3\texttt{</answer>}\texttt{<answer>}5\texttt{</answer>}\\\texttt{\textbf{<|im\_end|>}}\\
\texttt{\textbf{<|im\_start|>}user}\\\texttt{[Question]}\\\texttt{\{question\}}\\\\\texttt{[Assistant 1’s Answer]}\\\texttt{\{answer\_1\}}\\\\\texttt{[Assistant 2’s Answer]}\\\texttt{\{answer\_2\}}\\\texttt{\textbf{<|im\_end|>}}\\\texttt{\textbf{{<|im\_start|>}}}assistant\\\texttt{<think>}
\end{tcolorbox}
\vspace{-2mm}
\caption{System Prompt for RL Training}
\label{fig:system_prompt}
\vspace{-4mm}
\end{figure*}

\subsection{Judge Task Definition}
\label{sec:judge_task_definition}
We consider a judge task where, given a query $Q$ and two responses $A_1$ and $A_2$, the model evaluates them on a scale of 1–10 $(s_1, s_2)$, following the JudgeLM~\citep{zhu2025judgelm} setting. Evaluation criteria include helpfulness, relevance, accuracy, and level of detail, as specified in Prompt~\ref{fig:system_prompt}. The goal is to predict scores that align with human judgments, which determine whether $A_1$ wins, ties with, or loses to $A_2$.

\subsection{Reward Design for Judge Task}
\label{sec:reward_design}
Rule-based rewards have shown strong performance and are widely adopted~\citep{deepseekai2025deepseekr1incentivizingreasoningcapability}. For judge tasks involving reasoning, we design a reward function that combines structural and content-based components. Specifically, the reward $r$ for sample $i$ is defined as $r_i = \mathcal{R}_{\text{struct}} + \mathcal{R}_{\text{content}}$, inspired by~\cite{xie2025logicrl}.
\vspace{-3mm}
\paragraph{Structural Reward}
As shown in Figure~\ref{fig:system_prompt}, the structural reward $\mathcal{R}_{\text{struct}}$ ensures that the model output includes a structured reasoning process enclosed in \texttt{<think>...</think>} tags and final judgment scores $(s_1, s_2)$ enclosed in \texttt{<answer>...</answer>} tags, where $s_1, s_2 \in \{1, \dots, 10\}$. The reward components are:
\vspace{-2mm}
\begin{equation}
\mathcal{R}_{\text{format}} = 
\begin{cases}
1.0, & \text{if all tags are correct and in proper order} \\
-0.5, & \text{if } s_1, s_2 \notin \{1, \dots, 10\} \\
-1.0, & \text{if severe formatting errors exist}
\end{cases}
\vspace{-4mm}
\end{equation}

\paragraph{Content Reward}
The content reward $\mathcal{R}_{\text{content}}$ evaluates the accuracy and confidence of the predicted scores $(s_1, s_2)$ against the ground-truth labels $(s_1^*, s_2^*)$. It consists of three parts:

\textit{Relation Reward} in Eq. \ref{eq:2} ensures the model correctly ranks responses, aligning with the core goal of \textbf{comparative evaluation}. Aligning relative order is more important than absolute score matching in evaluation tasks, because win-loss relationships are more concerned, so the value is dominant in the content reward.
\vspace{-2mm}
\begin{equation}
r_{\text{relation}} = 
\begin{cases}
2.0, & \text{if } \text{sgn}(s_1 - s_2) = \text{sgn}(s_1^* - s_2^*) \\
-1.5, & \text{otherwise}
\end{cases}
\label{eq:2}
\vspace{-2mm}
\end{equation}

\textit{Absolute Reward} in Eq. \ref{eq:3} refines score-level accuracy, tolerating minor errors while penalizing large deviations. The scores are gradually refined by hierarchical rewards (complete match $\rightarrow$ partial match $\rightarrow$ no reward) to solve ‘\textbf{sparse rewards}’ in reinforcement learning: if relying only on relational rewards, the model may learn only coarse sequential judgements and ignore score calibration.
\vspace{-2mm}
\begin{equation}
r_{\text{absolute}} = 
\begin{cases}
1.0, & \text{if } |s_1 - s_1^*| + |s_2 - s_2^*| = 0 \\
0.6, & \text{if } r_{\text{relation}} = 2 \text{ and } |s_1 - s_1^*| + |s_2 - s_2^*| \leq 2 \\
0, & \text{otherwise}
\end{cases}
\label{eq:3}
\vspace{-3.5mm}
\end{equation}

\textit{Confidence Reward} in Eq. \ref{eq:4} promotes decisive judgments when the ranking is correct, avoiding ambiguous scores. \textbf{Harder judge problems result in ambiguous scores}, so we stimulate high confidence score to \textbf{improve ‘decision confidence’}, which results in more polarised decision output. We activate it only when the relative order is correct to avoid the model blindly expanding the variance.
\vspace{-2mm}
\begin{equation}
    r_{\text{confidence}} = 
\begin{cases}
0.2, & \text{if } r_{\text{relation}} = 2 \text{ and } |s_1 - s_2| \geq |s_1^* - s_2^*| \\
0, & \text{otherwise}
\end{cases}
\label{eq:4}
\vspace{-3mm}
\end{equation}

The final reward $r(s_1,s_2)$ is computed as:
\vspace{-1mm}
\begin{equation}
    r(s_1,s_2) = \underbrace{\mathcal{R}_{\text{format}}}_{\mathcal{R}_{\text{struct}}} + \underbrace{r_{\text{relation}} + r_{\text{absolute}} + r_{\text{confidence}}}_{\mathcal{R}_{\text{content}}}
    \vspace{-3mm}
\end{equation}

The design of structural and content rewards encourages the model to learn \textbf{distinguishing} good/bad and make relative judgments, \textbf{rather than pursuing absolute high rewards}, which is neglected by previous SFT methods on judges \citep{wang_pandalm_2024,zhu2025judgelm}. We discuss the sensitiveness to reward exact values and the relative reward relationship in RL rule-based training in Appendix \ref{app:rewardanalysis}.

\vspace{-1mm}
\subsection{RL Training Algorithm}

We utilize GRPO~\citep{deepseekai2025deepseekr1incentivizingreasoningcapability} as policy gradient algorithm, which eliminates the need for a separate value model in Proximal Proximal Policy Optimization (PPO)~\citep{schulman2017ppo} by grouping samples and normalizing \textbf{intra-group}
advantages for more practical training. Moreover, it reduce the data collection cost and increases generalizability than SFT because it does not need explicit reasoning chain to learn.

For policy \(\pi_\theta\) and reference model \(\pi_{\text{ref}}\), we define:

\paragraph{Group-wise Advantage Calculation}
For each query group $G_Q$ with its associated judgments $G_Q = \{(A_1, A_2, s_1, s_2)\}$, the mean $\mu_Q$ and the standard variation $\sigma_Q$ of reward $r$ in group $Q$ is calcuated as:
\vspace{-1mm}
\begin{equation}
    \mu_Q = \mathbb{E}_{(A_1,A_2) \sim G_Q}[r(s_1,s_2)], \quad 
\sigma_Q = \sqrt{\mathbb{E}_{(A_1,A_2) \sim G_Q}[(r(s_1,s_2)-\mu_Q)^2]}
\vspace{-3mm}
\end{equation}

The normalized advantage $\mathcal{A}$ for sample $i$ is defined for the optimization below to quantify how good $i$ is relative to the average action:
\vspace{-1mm}
\begin{equation}
    \mathcal{A}_i(s_1,s_2|Q) = \frac{r_i(s_1,s_2) - \mu_Q}{\sigma_Q + \eta}
\vspace{-1mm}
\end{equation}

In which $\eta$ is a smoothing term for numerical stability during training, preventing $\sigma_Q$ from being too small during normalization \citep{schulman2017ppo}. By normalizing intra-group advantages $\mathcal{A}$,  variance is reduced to improve the \textbf{stability} of policy updates. This allows the model to receive an effective learning signal even on difficult tasks where rewards are generally low, thereby \textbf{alleviating the training imbalance} caused by varying task difficulties across domains in Figure \ref{fig:judge_sftvsbase}.

\paragraph{Policy Optimization Objective}
The policy \(\pi_\theta\) of parameter $\theta$ is optimized using the following objective:
\vspace{-1mm}
\begin{equation}
\begin{split}
J_{\text{GRPO}}(\theta) = &\mathbb{E}_{Q \sim \mathcal{D}} \mathbb{E}_{(s_1,s_2) \sim \pi_\theta} \Bigg[ 
\min\Bigg( \frac{\pi_\theta(s_1,s_2|Q,A_1,A_2)}{\pi_{\text{old}}(s_1,s_2|Q,A_1,A_2)}\mathcal{A}_i(s_1,s_2|Q), \\
& \text{clip}\Big( \frac{\pi_\theta(s_1,s_2|Q,A_1,A_2)}{\pi_{\text{old}}(s_1,s_2|Q,A_1,A_2)}, 1-\epsilon, 1+\epsilon \Big)\mathcal{A}_i(s_1,s_2|Q) \Bigg) - \beta D_{\text{KL}}(\pi_\theta \| \pi_{\text{ref}}) \Bigg]
\end{split}
\label{eq:8}
\end{equation}
In which the clip range $\epsilon$ limits the magnitude of policy updates to prevent training instability, and the KL divergence is set to prevent $\pi_\theta$ from excessively deviating from reference model $\pi_{\text{ref}}$ with factor $\beta$ \citep{schulman2017ppo}.

\section{Experiments}

We empirically evaluate JudgeLRM to address the following research questions:

\begin{itemize}[leftmargin=*] 
    \item\textbf{Effectiveness} : How does JudgeLRM perform compared to other SFT and RL baselines (Section~\ref{sec:otherbasline}) and state-of-the-art LRMs and specialized judgement models (Section~\ref{sec:judge_perf})? 
    \item \textbf{Ablation Study} (Section~\ref{sec:ablation}): What are the individual contributions of different reward components to JudgeLRM’s performance? 
    \item \textbf{Reliability} (Section~\ref{sec:reliability}): Can JudgeLRM provide consistent and accurate comparative judgments? 
    \item \textbf{Case Study} (Section~\ref{sec:case_study}): In what ways does JudgeLRM demonstrate effective reasoning to support high-quality judgments? \end{itemize}

\subsection{Experimental Setup}
\label{sec:experiment_settings}

Our experiments mainly focus on pair-wise judge scenarios (since we also support other scenario as discussed in Section \ref{sec:judge_perf}), targeting on open-ended QA evaluation rather than closed-domain or expert-level tasks, and JudgeLM \citep{zhu2025judgelm}
PandaLM \citep{wang_pandalm_2024}, aligning directly with our objective. By contrast, benchmarks such as RewardBench~\citep{lambert2024rewardbenchevaluatingrewardmodels}, JudgeBench~\citep{tan2025judgebench}, while valuable, primarily address single-judge or domain-specific evaluation scenarios that fall outside the scope of this study. Our design emphasizes isolating the contribution of reasoning, rather than data scaling or architectural engineering, by assessing how RL enhances the generalization of judges over SFT, under the same dataset and base-model conditions as JudgeLM. Following prior work~\citep{zheng2024judging}, we prioritize controlled experiments on well-established benchmarks to ensure comparability.

\noindent\textbf{Datasets.}  
We evaluate on two established benchmarks.  
JudgeLM uses GPT-4 annotations as gold labels, comprising 100K training instances and a 5K test set, following the task formulation described in Section~\ref{sec:judge_task_definition}.  
PandaLM provides human-annotated pairwise comparisons with win/tie/loss labels on a 1K test set, offering complementary supervision. Together, these datasets allow us to assess performance across both GPT-aligned and human-grounded evaluation settings.

\noindent\textbf{Baselines.} We compare JudgeLRM against three categories of baselines:  
(1) \emph{Base, SFT and RL methods}: Base (Table \ref{tab:judgeneedreasoning}), SFT (Table \ref{tab:judgeneedreasoning},\ref{tab:rlcompare}), Bradley-Terry \citep{bradleyterry1952rank}, Direct Preference Optimization (DPO) \citep{rafailov2023dpo}, etc (Table \ref{tab:rlcompare}).
(2) State-of-the-art proprietary LLMs, including GPT-3.5, GPT-4, and Deepseek-R1 (Table \ref{tab:performance});
(3) \emph{Specialized judgment models}: Auto-J-13B, JudgeLM-7B/13B/33B, and PandaLM-70B (Table \ref{tab:performance}).
This setup enables a fair comparison across different alternatives, isolating the effect of reasoning-oriented training.

\noindent\textbf{Evaluation Metrics.} Following prior work, we evaluate model performance using agreement-based metrics: precision, recall, and F1 score,which quantify alignment with teacher model judgments.

\noindent\textbf{Implementation Details.} Our models, JudgeLRM-3B/7B/14B and JudgeLRM-4B/8B, are based on Qwen2.5-Instruct and Qwen3, respectively. All models including baselines are trained exclusively using reinforcement learning on the JudgeLM training set. The training is conducted over a single epoch, with a maximum prompt length of 1024 tokens and a maximum response length of 2048 tokens, using a batch size of 16. JudgeLRM-3B/4B is trained on 4×A100 80GB GPUs with a learning rate of 3e-7, while JudgeLRM-7B/8B/14B is trained on 8×A100 80GB GPUs with a learning rate of 1e-6. Following \citet{xie2025logicrl}\footnote{Realization \citep{xie2025logicrl} of Equation \ref{eq:8} \citep{deepseekai2025deepseekr1incentivizingreasoningcapability} does not include response-level length bias \citep{liu2025understandingr1zeroliketrainingcritical} in origin GRPO \citep{shao2024deepseekmathpushinglimitsmathematical}.}, the group size of $Q$ is set to 8, the clip range $\epsilon$ is set to 0.5, the KL divergence factor $\beta$ is set to 0.001,  and smooth term $\eta$ is set to 1e-6.

\subsection{Baselines: Additional SFT and RL Variants}
\label{sec:otherbasline}

\begin{table}[h]
\begin{center}
\resizebox{0.7\textwidth}{!}{
\begin{tabular}{lcccc}
\toprule
\multicolumn{1}{l}{Dataset} & \multicolumn{4}{c}{\textit{PandaLM (Human as ground truth. Out of distribution.)}} \\
\cmidrule(r){2-5}
\multirow{1}{*}{\textbf{Criteria}} & \multicolumn{1}{c}{\textbf{Agreement}} & \multicolumn{1}{c}{\textbf{Precision}} & \multicolumn{1}{c}{\textbf{Recall}} & \multicolumn{1}{c}{\textbf{F1}} \\

\hline \multicolumn{5}{l}{\textit{Method Baselines. (trained JudgeLM)}} \\
\hdashline
\quad Qwen2.5-3B-Instruct  & 68.50 & 50.92 & 56.13 & 51.57 \\
\quad Qwen2.5-3B-Instruct-Judge-SFT  & 70.57 & 67.09 & 73.36 & 66.10 \\
\quad DPO-Answer (Qwen2.5-3B-Instruct) & 67.27 & 44.79 & 50.07 & 47.27 \\

\quad CLS-RM (Qwen2.5-3B) & 58.15 & 51.70 & 51.69 & 51.69 \\

\quad Bradley-Terry (Qwen2.5-3B) & 58.94 & 59.04 & 59.06 & 58.94 \\

\quad SFT-Think (Qwen2.5-3B-Instruct) & 72.49 & 81.69 & 56.01 & 55.03 \\

\quad DPO-RC (Qwen2.5-3B-Instruct) & 68.67 & 64.54 & 69.43 & 65.23\\

\hdashline
\multicolumn{5}{l}{\textit{Ours.}} \\
\quad JudgeLRM-3B & 77.68 & 74.26 & 70.86 & 72.12\\

\bottomrule
\end{tabular}
}
\end{center}
\captionsetup{type=table}

\caption{Comparison with other RL and SFT methods.}

\label{tab:rlcompare}
\end{table}

To provide a comprehensive comparison, we construct several additional baselines using the Qwen2.5-3B family (base model in parentheses). These methods reflect alternative SFT or RL training paradigms but differ from our JudgeLRM design in how they handle preferences and reasoning.

\noindent\textbf{(1) DPO-Answer (Qwen2.5-3B-Instruct).}  
Optimizes a generative policy via DPO on chosen/rejected pairs using only win/loss signals, predicting preferences by comparing response likelihoods, focusing on generation preference rather than explicit judgment.

\noindent\textbf{(2) CLS-RM (Qwen2.5-3B).}  
Fine-tunes a reward model with a classification head. The RM assigns absolute scores to individual responses, and preferences are inferred by comparing these scores. Unlike JudgeLRM, this approach does not leverage relational reasoning across responses.

\noindent\textbf{(3) Bradley-Terry (Qwen2.5-3B).}  
Implements a pairwise preference model following the Bradley-Terry formulation, trained on chosen/rejected pairs with associated preference scores. The model predicts a scalar reward for each response, which is used for pairwise evaluation.

\noindent\textbf{(4) SFT-Think (Qwen2.5-3B-Instruct).}  
SFT using a JudgeLRM-style template where the model generates an explanation enclosed in \texttt{<think>} tags before producing its judgment. This setting isolates SFT under structural constraints. We find explanations fail to function as effective reasoning chains, leading to weaker adherence to structure and underperformance compared to SFT baselines.

\noindent\textbf{(5) DPO-RC (Qwen2.5-3B-Instruct).}  
Additional RL baseline following SPIN~\citep{chen2024spin}, trained with a content-based reward ($\mathcal{R}_{\text{content}}$) while adopting the JudgeLRM-style data template. This method performs competitively with SFT-trained models but still lags behind JudgeLRM-3B.

Overall, (1-3) remain limited because they abstract judgment into token-level preference modeling, losing relational understanding and the ability to provide rationale. (4) highlights SFT’s structural weakness, formatting instructions alone do not yield robust reasoning chains. (5) draws close to SFT baselines but fails to match the reasoning strength and structural fidelity of JudgeLRM.

\subsection{Judge Performance}
\label{sec:judge_perf} 

\begin{table}[!t]
\begin{center}
\resizebox{\textwidth}{!}{
\begin{tabular}{lcccccccc}
\toprule
\multicolumn{1}{l}{Dataset} & \multicolumn{4}{c}{\textit{JudgeLM (GPT-4 as ground truth. In distribution.)}} & \multicolumn{4}{c}{\textit{PandaLM (Human as ground truth. Out of distribution.)}} \\
\cmidrule(r){2-5} \cmidrule(r){6-9}
\multirow{1}{*}{\textbf{Criteria}} & \multicolumn{1}{c}{\textbf{Agreement}} & \multicolumn{1}{c}{\textbf{Precision}} & \multicolumn{1}{c}{\textbf{Recall}} & \multicolumn{1}{c}{\textbf{F1}} & \multicolumn{1}{c}{\textbf{Agreement}} & \multicolumn{1}{c}{\textbf{Precision}} & \multicolumn{1}{c}{\textbf{Recall}} & \multicolumn{1}{c}{\textbf{F1}} \\
\hline \multicolumn{9}{l}{\textit{Baselines. (* from original paper)}}  \\ 
\quad GPT-3.5* & 73.83 & 70.70 & 52.80 & 52.85 & 62.96 & 61.95 & 63.59 & 58.20 \\
\quad GPT-4* &-&-&-&-& 66.47 & 66.20 & 68.15 & 61.80 \\
\quad PandaLM-7B & 68.61 & 40.75 & 38.82 & 39.41 & 59.26 & 57.28 & 59.23 & 54.56 \\
\quad Auto-J-13B* & 74.86 & 61.65 & 57.53 & 58.14 &-&-&-&-\\
\quad JudgeLM-7B & 81.11 & 69.67 & 78.39 & 72.21 & 65.07 & 66.89 & 71.95 & 61.92\\
\quad JudgeLM-13B* & 84.33 & 73.69 & 80.51 & 76.17 &68.97& 68.21&74.15&65.12\\
\quad JudgeLM-33B* & \textbf{89.03} & 80.97 & 84.76 & 82.64 &75.18&69.30&\underline{74.93}&69.73\\
\quad PandaLM-70B* &-&-&-&- & 66.87 & 74.02 & 66.87 & 69.23\\

\quad Deepseek-R1 & - & - & - & - & 78.67 & 77.51 & 69.97 & 72.48 \\

\hdashline
\multicolumn{9}{l}{\textit{Ours. (Qwen3 Base)}} \\
\quad JudgeLRM-4B & 84.61 & \underline{86.82} & 84.24 & 85.51 & 77.88 & 72.33 & 73.67 & 72.87\\

\quad JudgeLRM-8B & \underline{86.69} & \textbf{89.16} & 85.76 & \textbf{87.43} & \underline{80.68} & 73.81 & 77.50 & \underline{75.26} \\
\quad JudgeLRM-4B-single & 81.17 & 80.13 & \underline{86.61} & 83.24 &  64.56 & 58.60 & 61.70 & 58.50\\
\quad JudgeLRM-8B-single & 83.20 & 81.64 & \textbf{88.84} & 85.09 & 71.27 & 65.07 & 69.34 & 65.30\\
\hdashline
\multicolumn{9}{l}{\textit{Ours. (Qwen2.5 Base)}} \\
\quad JudgeLRM-3B & 83.72 & 86.31 & 82.98 & 84.61 & 77.68 & 74.26 & 70.86 & 72.12\\
\quad JudgeLRM-7B & 83.74 & 85.84 & 83.65 & 84.73 & 78.28 & \underline{74.90} & \textbf{75.74} & 75.05\\
\quad JudgeLRM-14B & 85.25 & 86.64 & 85.91 & \underline{86.27} & \textbf{81.18} & \textbf{78.39} & 74.80 & \textbf{76.29} \\
\bottomrule
\end{tabular}
}
\end{center}
\captionsetup{type=table}

\caption{Performance on JudgeLM and PandaLM. As pairwise comparisons rarely yield ties, we exclude tie cases ($\sim$10\% of the test set) for more interpretable evaluation, which JudgeLRM-7B's F1 reaches 83.47, all metrics on PandaLM surpass DeepSeek-R1 (see Table~\ref{tab:excludetie} for details). More analysis of results and performance visualization are presented in Appendix \ref{sec:moreanalysis}.}
\label{tab:performance}
\vspace{-4mm}
\end{table}
We benchmark JudgeLRM against a range of competitive models, with results summarized in Table~\ref{tab:performance}. At the 7B scale, JudgeLRM-7B substantially outperforms its SFT counterpart (Qwen2.5-7B-Instruct-Judge-SFT) as well as the task-specific JudgeLM-7B model. Remarkably, \textbf{JudgeLRM-3B/4B surpasses GPT-4 on the human-annotated PandaLM benchmark, and scaling further to JudgeLRM-7B/8B matches or even surpass the performance of DeepSeek-R1.} Moreover, we also showcase JudgeLRM's ability to judge single answer in "-single" setting in Table~\ref{tab:performance} with first testing score through prompt \ref{fig:single} and pairwise judge test like CLS-RM. It surpasses all baselines with the same size, showing the adaptability of JugdeLRM.

\begin{figure}[t]
    \centering
    \includegraphics[width=0.6\linewidth]{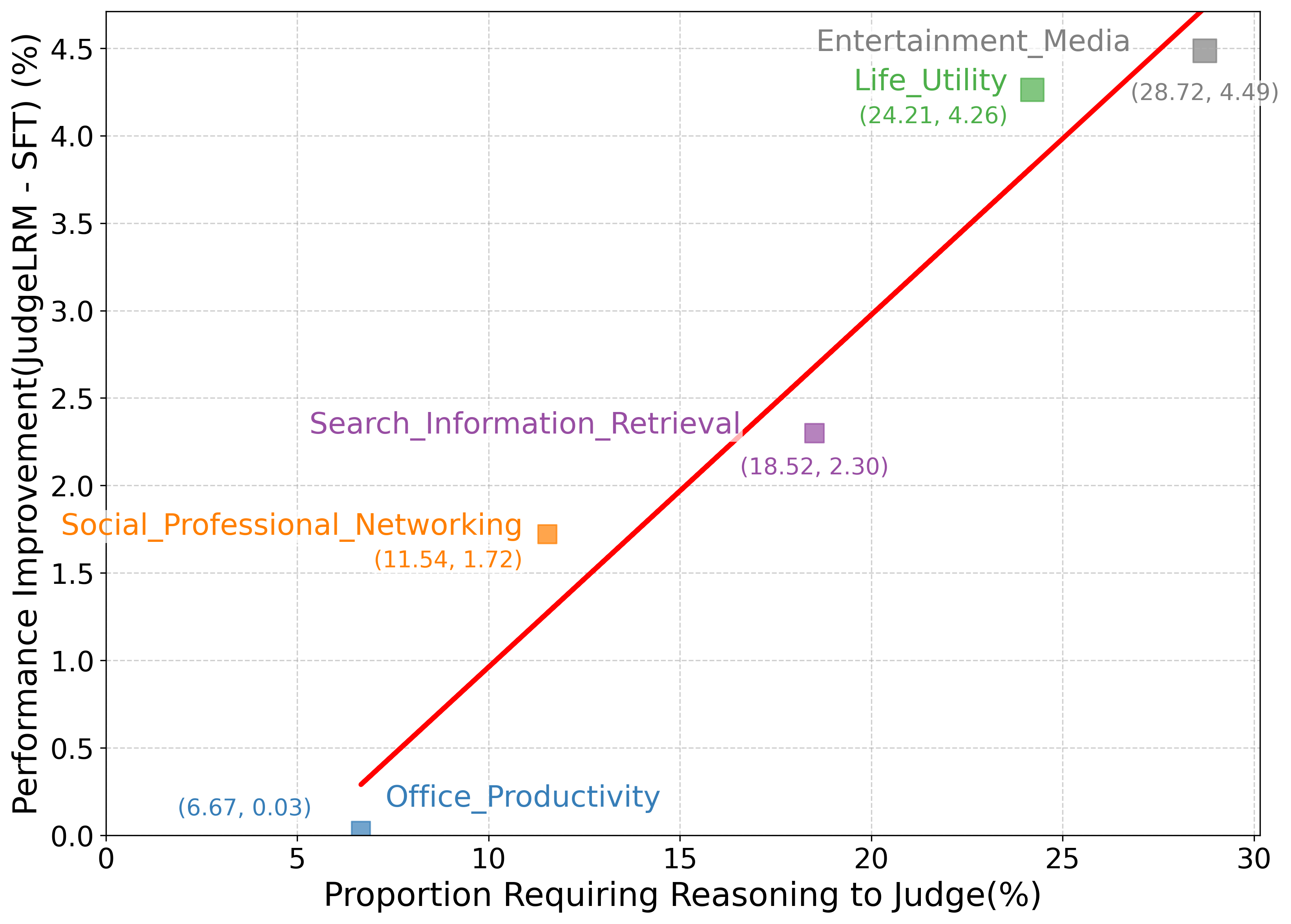}
\caption{Judgment performance improvement vs. reasoning requirement across domains. The Y-axis indicates the F1 score improvement of JudgeLRM-7B over the Qwen2.5-7B-Instruct-Judge-SFT baseline; the X-axis represents the proportion of tasks within each domain that require reasoning. Each point corresponds to a domain. A negative linear trend ($y = 0.2x - 1.05$, $R^2 = 0.95$) suggests that domains with a higher proportion of reasoning-intensive judge tasks see greater performance gains from JudgeLRM-7B. See Appendix \ref{sec:imp} for further analysis. } \vspace{-3mm}   \label{fig:judge_ourvssft}
\end{figure}

To investigate the source of these gains, we examine JudgeLRM-7B’s behavior on reasoning-intensive subsets of PandaLM. Figure~\ref{fig:judge_ourvssft} shows a positive correlation between performance improvements and the proportion of reasoning-demanding instances. Specifically, when comparing JudgeLRM-7B against its base model (Qwen2.5-7B-Instruct), we observe a correlation coefficient of $0.20$ between relative improvement and reasoning rate, with larger gains concentrated in reasoning-heavy categories. This trend provides direct evidence that judgment is inherently a reasoning-intensive task, and that reinforcement learning with outcome-driven rewards equips JudgeLRM with stronger evaluative reasoning than SFT-based baselines.

\subsection{Ablation Study} \label{sec:ablation}
To disentangle the effect of different reward components, we perform an ablation study on the human-annotated PandaLM benchmark. Our analysis focuses on the content-based rewards introduced in Section~\ref{sec:reward_design}, namely $r_{\text{absolute}}$ and $r_{\text{confidence}}$. These rewards explicitly encourage accurate scoring and calibrated confidence, thereby guiding models to verify their reasoning chains and revise errors within the \texttt{<think>} step before finalizing a judgment. 

As shown in Table~\ref{tab:ablation}, removing content rewards leads to a consistent 2–5\% drop in F1 score. Without these signals, models tend to produce superficial explanations or fail to detect inconsistencies between evidence and verdict, underscoring that judgment accuracy requires more than structural formatting alone. It demonstrates that outcome-driven content rewards are crucial for eliciting faithful reasoning, effective error correction, and reliable decision-making in JudgeLRM.

\begin{table}[h]
\begin{center}
\resizebox{0.7\textwidth}{!}{
\begin{tabular}{lcccc}
\toprule
\multirow{1}{*}{Test on PandaLM} & \multicolumn{1}{c}{\textbf{Agreement}} & \multicolumn{1}{c}{\textbf{Precision}} & \multicolumn{1}{c}{\textbf{Recall}} & \multicolumn{1}{c}{\textbf{F1}} \\
\hline
\quad JudgeLRM-7B & \textbf{78.28} & 74.90 & \textbf{75.74} & \textbf{75.05}\\
\quad\quad - w/o. $r_{\text{abs}}+r_{\text{conf}}$ & 75.78 & 69.09 & 73.69 & 70.36 \\
\quad\quad - w. $\mathcal{R}_{\text{length}}$ & \textbf{78.28} & \textbf{75.81} & 69.19 & 71.34 \\
\bottomrule
\end{tabular}
}
\end{center}
\captionsetup{type=table}

\caption{Ablation study of 7B models on the human-annotated PandaLM dataset.}
\vspace{-4mm}
\label{tab:ablation}
\end{table}

\paragraph{Effect of Incentivizing Lengthy response} 

During JudgeLRM training, we observe that both reasoning and response lengths naturally increase with training steps, with larger models (e.g., 7B) producing longer outputs than smaller ones (3B) (Appendix, Figure~\ref{fig:step}). To test whether explicitly encouraging longer reasoning chains improves performance, we introduced the following length reward:  
\vspace{-2mm}

\begin{equation}
\mathcal{R}_{\text{length}} = 
\begin{cases}
0.2, & \text{if the reasoning chain exceeds 120 tokens} \\
-1.0, & \text{if the maximum token limit is reached} \\
0, & \text{otherwise}
\end{cases}
\end{equation}

However, as shown in Table \ref{tab:ablation}, simply incentivizing slightly longer answers degraded results of about 3\%, suggesting that merely increasing output length (potentially via repetition) \textbf{does not benefit} the judge task.

\subsection{Reliability of Model Judgment} \label{sec:reliability}

A key property of judgment models is reliability -- the ability to produce consistent and accurate judgments regardless of the order in which candidate answers are presented. To evaluate the reliability of JudgeLRM and representative baselines, we conduct experiments on the JudgeLM dataset by permuting the order of answers. We assess three aspects: (1) self-consistency across permutations, (2) positional bias toward the first or second answer, and (3) the bias gap ($\Delta_{\text{bias}}$), which captures variance in position-based preferences.

As shown in Table~\ref{tab:bias}, JudgeLRM substantially improves self-consistency compared to both its base Instruct model and the task-specific JudgeLM baseline. Additionally, JudgeLRM significantly reduces bias toward the first answer while maintaining comparable bias levels toward the second, achieving more balanced and robust evaluation. It demonstrate that reinforcement learning with outcome-driven rewards not only enhances reasoning ability but also mitigates order sensitivity.

\begin{table}[t]
\centering
\resizebox{0.7\textwidth}{!}{ 
\begin{tabular}{lccccc}

\toprule
\multirow{2}{*}{Methods} & Consistency $\uparrow$ & Bias $\downarrow$ & Bias $\downarrow$ & \multirow{2}{*}{Delta Bias $\downarrow$} \\
 & (w/ swap.) & toward 1st & toward 2nd & \\
\midrule
JudgeLM score-first* & 73.45 & 19.83 & 6.72 & 13.11\\
GPT-4* & 85.82 & 5.55 & 3.09 & 2.46 \\
\midrule
Qwen-7B-Instruct 0shot &  77.11 & 12.96 & 9.93 & 3.04 \\
JudgeLRM-7B &  84.50 & 5.39 & 10.11 & 4.72 \\
JudgeLRM-8B & 89.55 & 4.04 & 6.40 & 2.36\\
\bottomrule
\end{tabular}
}
\caption{Assessment of position bias on the \textit{val} split of JudgeLM. We evaluate model self-consistency under different answer order permutations, quantify bias toward the first or second answer, and report the gap between these biases ($\Delta_{\text{bias}}$).}
\label{tab:bias}
\vspace{-4mm}
\end{table}

\subsection{Case Study} \label{sec:case_study}

JudgeLRM exhibits superior judgment by learning to reason explicitly. To probe how it produces informed, high-quality decisions, we analyze its responses for emergent reasoning behaviors akin to the LRM-style cognitive patterns reported for DeepSeek-R1 \citep{gandhi2025cognitivebehavior}. We first illustrate these patterns in DeepSeek-R1 (Figure~\ref{fig:criteria_evaluation_r1_2}), then examine a representative JudgeLRM example (Table~\ref{tab:judgelrm-casestudy}) in which the model evaluates two study notes and selects the one that better supports learning U.S. history.

The JudgeLRM response demonstrates several hallmark reasoning strategies characteristic of high-quality evaluators:
\vspace{-1mm}

\begin{itemize}[leftmargin=*]
\item \textbf{Verification:} Checking claims against the prompt and available evidence to avoid misinformation and support reliable judgments.
\item \textbf{Subgoal setting:} Decomposing the evaluation into concrete, interpretable criteria (coverage, chronology, specificity), yielding a structured assessment.
\item \textbf{Double-checking and reflection:} Re-examining intermediate conclusions to reduce errors and reconcile inconsistencies before finalizing the verdict.
\item \textbf{Error identification:} Spotting factual or logical flaws in candidate responses and localizing them to specific spans.
\item \textbf{Decision justification:} Articulating a clear, criterion-linked rationale that explains \emph{why} the preferred answer better satisfies the task.
\end{itemize}

We further observe a consistent three-stage structure: evaluate Assistant~1, evaluate Assistant~2, then synthesize a comparative assessment, mirroring human-like evaluation behavior \citep{dasgupta2024languagemodelshumanlikecontent}. This structure reflects integration of semantic understanding, logical inference, and hierarchical comparison. Notably, the judge-wise reward design jointly incentivizes \emph{structural coherence} (well-formed \texttt{<think>} and answer sections) and \emph{content accuracy} (relation, absolute, and confidence signals), reinforcing these behaviors during training.

\begin{table}[h]
\centering
\fontsize{9}{10}\selectfont
\begin{tabular}{p{14cm}}
\toprule
5,5. Both responses were relevant and accurate. They both provided a detailed overview of the history of the United States, including important events and dates. However, the response was repetitive, mentioning the same events twice, which is why it didn't receive a perfect score.\\
\\
\hline
\end{tabular}
\caption{Qwen2.5-7B-Instruct-Judge-SFT fails to response the question in Table \ref{tab:judgelrm-casestudy} (ID 549).}
\vspace{-3mm}
\label{tab:QWEN-casestudy}
\end{table}

\vspace{-1mm}
\section{Conclusion}
Our work demonstrates that judgment tasks for LLMs are inherently reasoning-intensive, with SFT-trained models struggling in high-reasoning domains. By reframing evaluation as an RL problem with judge-specific, outcome-driven rewards, JudgeLRM learns to generate structured, trustworthy reasoning paths. Empirical results show that JudgeLRM not only outperforms leading models like GPT-4 and DeepSeek-R1, but also scales effectively across model sizes from 3B to 14B. Beyond performance, our analysis reveals that successful judgment involves complex reasoning behaviors like verification, sub-goal planning, and justification, highlighting the need to treat judgment not as mere scoring, but as a process of structured reasoning. We hope this paves the way for future research on rigorous, reliable LLM evaluators.

\begin{table}[h!]
\centering
\fontsize{9}{10}\selectfont
\begin{tabular}{p{14cm}}
\toprule
Question: \\
(Instruction:) Include important study notes and key points that someone should know about the given subject.
(Input:) history of the USA \\

[Assistant 1’s Answer]
The history of the United States is one of the most influential and influential countries in the world. It has had a profound impact on the world and continues to shape the world today. The United States has had a long and storied history, from its founding in 1775 to the present day. It has had a tumultuous and tumultuous history, from the Revolutionary War to the Civil War and beyond. The United States has had a tumultuous and tumultuous history, from the Revolutionary War to the Civil War and beyond. The United States has had a long and storied history, from its founding in 1775 to the present day. It has had a profound impact on the world and continues to shape the world today.

[Assistant 2’s Answer]
1. The United States of America was founded in 1776.
2. The Declaration of Independence was signed in 1776.
3. The Constitution of the United States of America was signed in 1787.
4. The Civil War began in 1861.
5. The Emancipation Proclamation was issued in 1863.
6. The 13th Amendment was ratified in 1865.
7. The 14th Amendment was ratified in 1868.
8. The 15th Amendment was ratified in 1870.
9. The 16th Amendment was ratified in 1913.
10. The 17th Amendment was ratified in 1913.
11. The 18th Amendment was ratified in 1919.
12. The 19th Amendment was ratified in 1920.
13. The 20th Amendment was ratified in 1933.
14. The 21st Amendment was ratified in 1933.\\
\midrule
Response: \texttt{<think>} \\
Assistant 1 provided a very broad, somewhat repetitive statement on the history of the United States which did not include specific key events, dates, or figures that would be considered important in the narrative of U.S. history. This kind of vague and repetitive phrasing might make it difficult for someone seeking specific historical information to gain an accurate or clear understanding. While it correctly noted the overall influence and significant historical periods like the Revolutionary War and the Civil War, this approach lacks specificity which can be misleading or uninformative.

\textcolor{red}{In contrast, Assistant 2's answer is quite accurate and relevant in terms of providing a list of critical events and amendments in U.S. history. Each point listed is accurate, such as the dates when key documents were signed, the years of major constitutional changes, and significant legislative acts like the Emancipation Proclamation and various constitutional amendments related to civil rights and representation.}\textcolor{blue}{\textit{\textbf{ (Verification)}}} \textcolor{red}{By including these specific details, the accuracy and relevance of the information provided by Assistant 2 is significantly higher, making it easier for someone to trace specific aspects of American history through these milestones.}\textcolor{blue}{\textit{\textbf{ (Subgoal Setting)}}}

\textcolor{red}{Accuracy is also a concern in Assistant 1's answer since the statement "from its founding in 1775" is inaccurate; the actual founding date recognized in the United States is July 4, 1776, with the signing of the Declaration of Independence, rather than 1775. }\textcolor{blue}{\textit{\textbf{ (Verification, Double Checking)}}}  \textcolor{red}{Additionally, the repetition might confuse readers, suggesting a lack of clarity or precision in how the history is presented, which could mislead people who are looking for a clear timeline or significant events.}\textcolor{blue}{\textit{\textbf{ (Reflection, Error Identification)}}}

\textcolor{red}{Both answers need to be more precise, but given the task, the importance lies in identifying key events, dates, and significant changes that shaped the nation.}\textcolor{blue}{\textit{\textbf{ (Subgoal Setting)}}} \textcolor{red}{Thus, the accuracy and relevance in listing specific, factual events and their exact timing in history would be a more appropriate way to address the question asked.}\textcolor{blue}{\textit{\textbf{ (Decision Justification)}}}

Given these considerations, while both answers are somewhat relevant to the history of the United States, the accuracy and level of detail in the information provided by Assistant 2 would likely be seen as more accurate and relevant because they pinpoint specific, significant events and changes in the country's development.\texttt{</think>}

\texttt{<answer>2</answer><answer>9</answer>}\\
\\
\hline
\end{tabular}
\caption{JudgeLRM-7B prediction on a PandaLM case (ID 549). Through reinforcement learning, JudgeLRM demonstrates diverse emergent reasoning behaviors -- capabilities highlighted in blue and the corresponding response segments in red -- even though such behaviors were not explicitly present in the training data. }
\label{tab:judgelrm-casestudy}
\vspace{-1mm}
\end{table}



\bibliography{colm2025_conference}
\bibliographystyle{iclr2026_conference}

\appendix
\section{Appendix}

\section*{Details of PandaLM Category Analysis}
\label{sec:pandalm_category}
\begin{table}[h!]
\centering
\renewcommand{\arraystretch}{1.2} %
\setlength{\tabcolsep}{4pt} 
\resizebox{\textwidth}{!}{ 
\begin{tabular}{lrrrrr}
\toprule
Metric & Entertainment\_Media & Office\_Productivity & Life\_Utility & Search\_Information\_Retrieval & Social\_Professional\_Networking \\
\midrule
Total & 195 & 105 & 190 & 108 & 104 \\
Reason (\%) & 28.72 & 6.67 & 24.21 & 18.52 & 11.54 \\
F1\_Qwen-7B-Instruct & 56.95 & 59.71 & 57.15 & 44.06 & 50.08 \\
F1\_Qwen-7B-Instruct-Judge-SFT & 62.52 & 74.88 & 60.75 & 58.11 & 58.43 \\
F1\_JudgeLRM-7B & 67.01 & 74.86 & 65.01 & 60.41 & 60.15 \\
\bottomrule
\end{tabular}
}
\caption{Model F1 performance comparison by categories.}
\label{tab:judge_reasoning_analysis}
\end{table}

The five main categories are collated from PandaLM "motivation\_app" label. Entertainment\_Media includes Netflix, IMDB, Spotify, YouTube, ESPN, Instagram, Facebook, Twitter, Telegram. Office\_Productivity is from MS Excel, MS Powerpoint, Google Sheet, Jira, Google Meet, Gmail. Life\_Utility is from Weather, Tasty, Sudoku, Goodreads, Yelp, traipadvisor.com, Redfin, Play Store, Amazon, Wysa, Real Estate. Search\_Information\_Retrieval is from Google Search, Quora, Reddit, CNN News, Indeed, Meetup Netflix, IMDB. Social\_Professional\_Networking is from LinkedIn, Messenger, Blogger. The “need reason” label is assigned according to the evaluation scale in Prompt \ref{fig:reasoning_eval_prompt}, and we define that scores 1-4 denote cases where reasoning is unnecessary, while scores 5-10 indicate that reasoning is needed. The results in Table \ref{tab:judge_reasoning_analysis}. We also showcase a subset of "motivation\_app" (query category) to demonstrate the improvement from SFT to JudgeLRM.

\begin{table}[h!]
\centering
\renewcommand{\arraystretch}{1.2} 
\setlength{\tabcolsep}{4pt} 
\resizebox{\textwidth}{!}{ 
\begin{tabular}{lccc}
\toprule
\multirow{1}{*}{Methods}
 & Wolframalpha35  & Grammarly30 & Gmail44 \\
\midrule
\multicolumn{4}{l}{\textit{Judge w/o reference (Ours).}} \\
Qwen-7B-Instruct & 45.71, 50.93, 53.21, 45.14 & 63.33, 54.55, 41.67, 46.96 & 61.36, 54.59, 69.44, 46.69 \\  
Qwen-7B-Instruct-Judge-SFT & 48.57, 46.01, 53.97, 46.27  & 73.33, 56.37, 46.30, 50.48 & 70.45, 60.56, 77.30, 55.60 \\
JudgeLRM-7B & 60.00, 58.97, 64.22, 58.33  & 76.67, 61.11, 50.93, 55.56 & 75.00, 60.43, 80.71, 59.49 \\
\bottomrule
\end{tabular}
}
\caption{Selected dataset results for PandaLM sub testset. Accuracy (Acc.), Precision (Prec.), Recall (Rec.), and F1-score (F1).}
\label{tab:domain_results}
\end{table}

\begin{figure*}[h]
\centering
\begin{tcolorbox}[colback=blue!5!white, colframe=blue!70!black, title=Prompt for Single Case Judgement, width=1.0\textwidth, sharp corners=south]
\texttt{\textbf{<|im\_start|>system}}\\You are a helpful assistant. The assistant first performs a detailed, step-by-step reasoning process in its mind and then provides the user with the answer. The reasoning process and answer are enclosed within \texttt{<think> </think>} and \texttt{<answer> </answer>} tags, respectively. Now the user asks you to judge ONE AI assistant's response to the question. Provide a single score from 1-10 (higher=better). Criteria include helpfulness, relevance, accuracy, and level of detail. Avoid bias from order, length, or style. After thinking, provide ONLY the numeric score within \texttt{<answer> </answer>} tags.\\\texttt{\textbf{<|im\_end|>}}\\
\texttt{\textbf{<|im\_start|>}user}\\\texttt{[Question]}\\\texttt{\{question\}}\\\\\texttt{[Assistant's Answer]}\\\{answer\}\\\\\texttt{\textbf{<|im\_end|>}}\\\texttt{\textbf{{<|im\_start|>}}}assistant\\\texttt{<think>}
\end{tcolorbox}
\caption{Prompt for single case judgement.}
\label{fig:single}
\end{figure*}

\begin{figure}[H] 
\centering 
\begin{tcolorbox}[
    colback=blue!5!white,     
    colframe=blue!70!black,   
    title=Prompt for Reasoning Ability Judgement, 
    width=1.0\textwidth,      
    sharp corners=south       
]
For the data provided below, "response1" and "response2" represent two responses generated for the given "instruction" and "input". Consider the task of judging the performance of "response1" and "response2" in response to the "instruction" and "input".\\\\
On a scale of 1 to 10, rate the level of reasoning ability needed to perform this judgment.\\ 
Please provide your response in EXACTLY the following format:\\
\texttt{----------------------------------------}\\
\texttt{Score: [your score, an integer between 1 and 10]}\\
\texttt{Explanation: [your explanation]}\\
\texttt{----------------------------------------}\\\\
\texttt{Instruction: {item['instruction']}}\\\\ 
\texttt{Input: {item['input']}}\\\\
\texttt{Response1: {item['response1']}}\\\\
\texttt{Response2: {item['response2']}}
\end{tcolorbox}
\caption{Prompt format for evaluating the reasoning difficulty of judging AI responses. Intended for appendix.} 
\label{fig:reasoning_eval_prompt} 
\end{figure}

\begin{figure}[h]
\centering
\begin{tcolorbox}[
    colback=blue!5!white,      
    colframe=blue!70!black,    
    title={Case Example 1 (idx: 145)}, 
    width=\textwidth,          
    sharp corners=south,       
    fontupper=\small,          
    boxrule=1pt,               
]
"idx": 145,\\[1ex]
"motivation\_app": "Wolfram alpha",\\[1ex]
"cmp\_key": "llama-7b\_opt-7b",\\[1ex]
"instruction": "Solve this equation.",\\[1ex]
"input":  \["x^3 - 4x^2 + 6x - 24 = 0"\],\\[1ex]
"response1": "x = 2",\\[1ex]
"response2": "x = 0",\\[1ex]
"annotator1": 0,\\[1ex]
"annotator2": 0,\\[1ex]
"annotator3": 0,\\[1ex]
"label": 0,\\[1ex]
"needed\_reasoning\_rate1-10": 3,\\[1ex]
\parbox[t]{\textwidth}{%
  "rate\_explanation": "The task involves checking whether the responses solve the given cubic equation. Verifying 'x = 2' or 'x = 0' requires substituting the values into the equation and performing basic arithmetic, which is straightforward. No advanced reasoning or complex problem-solving is needed, as the equation can be evaluated directly. However, recognizing that 'x = 2' is a correct solution and 'x = 0' is incorrect involves minimal reasoning."
}
\\[1ex]
"----------------------------------------" 
\end{tcolorbox}
\caption{Example Case 1 (idx: 145) data for reasoning evaluation.}
\label{fig:case_example_1}
\end{figure}

\begin{figure}[H] 
\centering 
\begin{tcolorbox}[
    colback=blue!5!white,     
    colframe=blue!70!black,   
    title=Case Example 2 (idx: 549), 
    width=1.0\textwidth,      
    sharp corners=south       
]
"idx": 549, \\\\
"motivation\_app": "Google Docs", \\\\ 
"cmp\_key": "cerebras-gpt-6.7B\_pythia-6.9b", \\\\ 
"instruction": "Include important study notes and key points that someone should know about the given subject.", \\\\
"input": "history of the USA", \\\\
"response1": "The history of the United States is one of the most influential and influential countries in the world. It has had a profound impact on the world and continues to shape the world today. The United States has had a long and storied history, from its founding in 1775 to the present day. It has had a tumultuous and tumultuous history, from the Revolutionary War to the Civil War and beyond. The United States has had a tumultuous and tumultuous history, from the Revolutionary War to the Civil War and beyond. The United States has had a long and storied history, from its founding in 1775 to the present day. It has had a profound impact on the world and continues to shape the world today.", \\\\ 
"response2": \\ 
   "1. The United States of America was founded in 1776." \\ 
   "2. The Declaration of Independence was signed in 1776." \\
   "3. The Constitution of the United States of America was signed in 1787." \\
   "4. The Civil War began in 1861." \\
   "5. The Emancipation Proclamation was issued in 1863." \\
   "6. The 13th Amendment was ratified in 1865." \\
   "7. The 14th Amendment was ratified in 1868." \\
   "8. The 15th Amendment was ratified in 1870." \\
   "9. The 16th Amendment was ratified in 1913." \\
   "10. The 17th Amendment was ratified in 1913." \\
   "11. The 18th Amendment was ratified in 1919." \\
   "12. The 19th Amendment was ratified in 1920." \\
   "13. The 20th Amendment was ratified in 1933." \\
   "14. The 21st Amendment was ratified in 1933." \\
"annotator1": 2, \\
"annotator2": 2, \\
"annotator3": 2, \\
"label": 2, \\
"needed\_reasoning\_rate1-10": 7, \\ 
"rate\_explanation": "The task requires evaluating the quality of responses based on their adherence to the instruction to include important study notes and key points about the history of the USA. Response1 is repetitive and lacks specific details, while Response2 provides a clear, concise list of key historical events. The reasoning needed to judge these responses involves assessing clarity, specificity, and relevance to the instruction, which is moderately complex." \\ 
"----------------------------------------" 
\end{tcolorbox}
\caption{Example Case 2 (idx: 549) data for reasoning evaluation.} 
\label{fig:case_example_2} 
\end{figure}

\begin{figure*}[!h]
\centering
\begin{tcolorbox}[colback=blue!5!white, colframe=blue!70!black, title=Prompt for JudgeLM, width=1.0\textwidth, sharp corners=south]
You are a helpful and precise assistant for checking the quality of the answer.\\[\baselineskip]\texttt{[Question]\\[\baselineskip]\{instruction\}\\\\\{input\}\\[\baselineskip][The Start of Assistant 1's Answer]\\[\baselineskip]\{response1\}\\[\baselineskip][The End of Assistant 1's Answer]\\[\baselineskip][The Start of Assistant 2's Answer]\\[\baselineskip]\{response2\}\\[\baselineskip][The End of Assistant 2's Answer]\\[\baselineskip][System]}\\[\baselineskip]We would like to request your feedback on the performance of two AI assistants in response to the user question displayed above.\\\\Please rate the helpfulness, relevance, accuracy, level of details of their responses. Each assistant receives an overall score on a scale of 1 to 10, where a higher score indicates better overall performance.\\\\Please first output a single line containing only two values indicating the scores for Assistant 1 and 2, respectively. The two scores are separated by a space. In the subsequent line, please provide a comprehensive explanation of your evaluation, avoiding any potential bias and ensuring that the order in which the responses were presented does not affect your judgment.\\\\[\baselineskip]\#\#\# Response:
\end{tcolorbox}
\caption{Prompt for JudgeLM.}
\label{fig:promptjudgelm}
\end{figure*}

\begin{figure*}[!h]
\centering
\begin{tcolorbox}[colback=blue!5!white, colframe=blue!70!black, title=Prompt for PandaLM, width=1.0\textwidth, sharp corners=south]
Below are two responses for a given task. The task is defined by the Instruction with an Input that provides further context. Evaluate the responses and generate a reference answer for the task.\\\\\#\#\# Instruction:\\\\\texttt{\{instruction\}}\\\\\#\#\# Input:\\\texttt{\{input\}}\\\\\#\#\# Response 1:\\\texttt{\{resp1\}}\\\\\#\#\# Response 2:\\\texttt{\{resp2\}}\\\\\#\#\# Evaluation:\\
\end{tcolorbox}
\caption{Prompt for PandaLM.}
\label{fig:promptpandalm}
\end{figure*}

\section*{Prompt for Accessing the Necessity of Reasoning When Judging }
\label{app:necessityreasoning}
We show prompt in Fig. \ref{fig:reasoning_eval_prompt} to rate the level of reasoning ability needed
to perform the judgment and two cases of rating in Fig. \ref{fig:case_example_1} and Fig. \ref{fig:case_example_2}. The reasoning rate is not totally decided by the requirement for reasoning to answer the question. In Fig. \ref{fig:case_example_1}, judging a math problem doesn't need reasoning. In Fig. \ref{fig:case_example_2}, judging a writing problem needs reasoning.

The “reasoning-required” scores in Figure \ref{fig:reasoning_eval_prompt} were generated by GPT-4. For scalability, we manually label 100 samples with ‘whether reasoning is required’ was associated with a GPT-4 score Kappa coefficient of 0.82.

\section*{Prompt for Baselines}
\label{app:prompt}
Fig. \ref{fig:system_prompt} shows the prompt for the inference of JudgeLRM. Fig. \ref{fig:single} shows the prompt for single case judgement. For reference, Fig. \ref{fig:promptjudgelm} shows the prompt for the inference of JudgeLM. Fig. \ref{fig:promptpandalm} shows the prompt for the inference of PandaLM. 

\section*{Further Reward Design Analysis}

\label{app:rewardanalysis}

In this section we discuss the sensitiveness to reward exact values and the relative reward relationship in RL rule-based training. RL rule-based training is insensitive to reward exact values under the setting of the paper. We reach it by slightly modifying the reward to Relation Reward (+1.1/-1.1 in Eq. \ref{eq:2}, Absolute Reward (+1.0/+0.5 in Eq. \ref{eq:3}, Confidence Reward (+0.4 in Eq. \ref{eq:4} without disrupting their relative order under 3B settings.

\begin{table}[h]
\begin{center}
\resizebox{0.7\textwidth}{!}{
\begin{tabular}{lcccc}
\toprule
\multirow{1}{*}{Test on PandaLM} & \multicolumn{1}{c}{\textbf{Agreement}} & \multicolumn{1}{c}{\textbf{Precision}} & \multicolumn{1}{c}{\textbf{Recall}} & \multicolumn{1}{c}{\textbf{F1}} \\
\hline
\quad Original reward & 77.68 & 74.26 & 70.86 & 72.12 \\
\quad Slightly modified & 77.65 & 74.21 & 70.90 & 72.10 \\
\bottomrule
\end{tabular}
}
\end{center}
\caption{Impact of reward function modification (JudgeLM-3B).}
\label{tab:reward1}
\end{table}
As the variance of each metric on multiple experiments is smaller than 0.04, the t-test shows p$\ge$ 0.05 on each metric, so slightly modifying the reward is insignificant on the performance.

It is possibly an advantage of GRPO training because GRPO grouping samples and normalizing intra-group advantages, which reduce variance and improve stability in strategy updates and encourage the model to learn distinguishing good/bad within problem groups and make relative judgments, rather than pursuing absolute high rewards.

However, if we changes the relative reward relationship, the performance decreases. We change the reward relationship to Relation Reward (+1.0/-1.5 in Eq. \ref{eq:2}, Absolute Reward (+2.0/+1.5 in Eq. \ref{eq:3}), Confidence Reward (+1.8 in Eq. \ref{eq:4}) under 3B settings, and the results are shown below.

\begin{table}[h]
\begin{center}
\resizebox{0.7\textwidth}{!}{
\begin{tabular}{lcccc}
\toprule
\multirow{1}{*}{Test on PandaLM} & \multicolumn{1}{c}{\textbf{Agreement}} & \multicolumn{1}{c}{\textbf{Precision}} & \multicolumn{1}{c}{\textbf{Recall}} & \multicolumn{1}{c}{\textbf{F1}} \\
\hline
\quad Original reward & 77.68 & 74.26 & 70.86 & 72.12 \\
\quad Changed relative & 75.48 & 68.41 & 70.87 & 69.31 \\
\bottomrule
\end{tabular}
}
\end{center}
\caption{Impact of relative relationship change (JudgeLM-3B).}
\label{tab:reward2}
\end{table}

It shows that the relative reward relationship is sensitive to the performance.

\section*{Improvements by different categories}
\label{sec:imp}
\begin{figure}[h]
    \centering
    \includegraphics[width=0.65\linewidth]{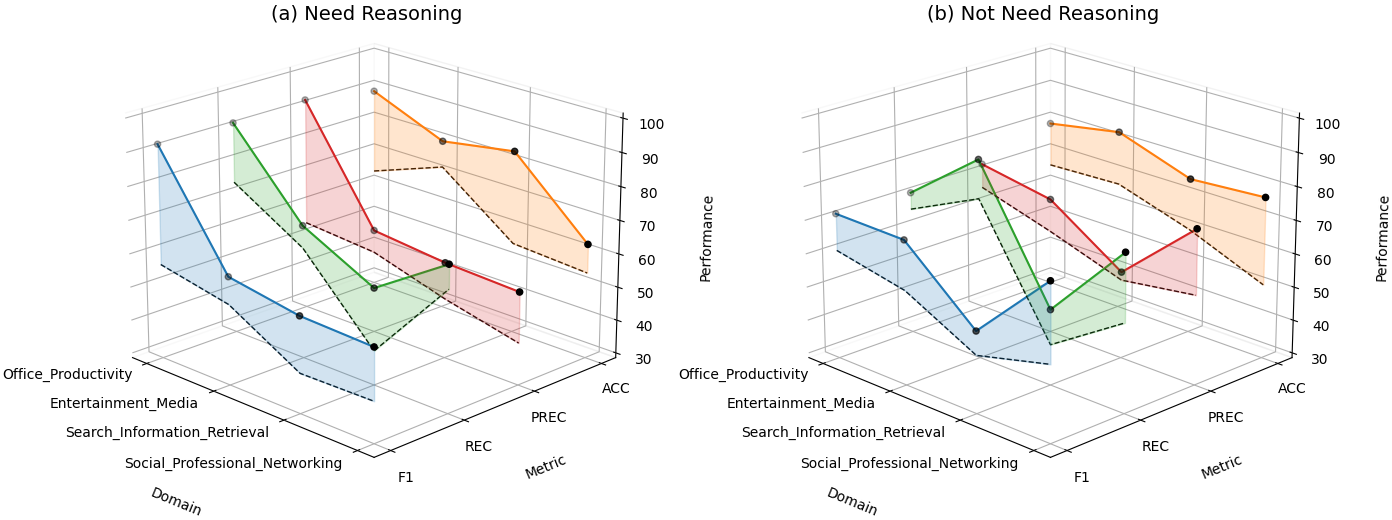}
\caption{Improvements from different categories.}    \label{fig:3dpic}
\end{figure}
Fig. \ref{fig:3dpic} shows improvements the JudgeLRM-7B's average F1 gain over SFT and that the gains persist even in reasoning‑heavy categories, showcasing the necessity of RL for judges.

\section*{More Performance Analysis PandaLM testset}
\label{sec:moreanalysis}
\begin{figure}
    \centering
    \includegraphics[width=0.7\linewidth]{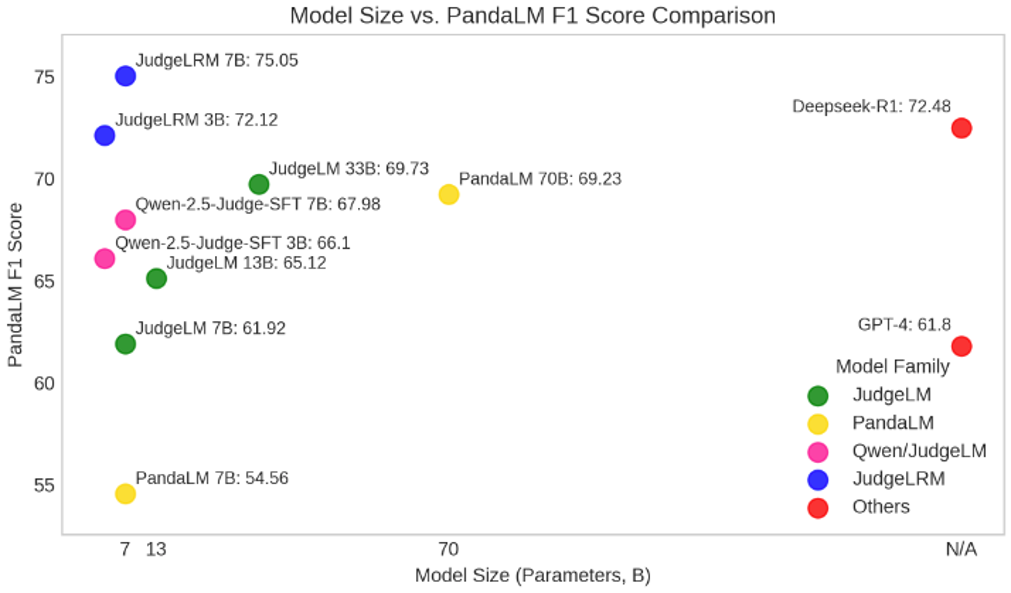}
    \caption{Visualization of the performance.}
    \label{figure:visual}
\end{figure}

\textbf{Visualization}
Figure \ref{figure:visual} presents a visalization of performance, proving the carefully crafted judge-wise rewards enable the model to learn reasoning effectively.

\textbf{Performance on PandaLM testset excluding ties.}

Since scoring pair-wise does not easily yield a tie, we simultaneously give the scoring of the pandalm for which the test set excludes ties (10\% in total) and treats the predicted tie as choosing the first assistant. The result is shown in Table \ref{tab:excludetie}.

\begin{table}[h]
\begin{center}
\resizebox{0.9\textwidth}{!}{
\begin{tabular}{lcccc}
\toprule
\multicolumn{1}{l}{Dataset} & \multicolumn{4}{c}{\textit{PandaLM (Human as ground truth)}} \\
\cmidrule(r){2-5}
\multirow{1}{*}{\textbf{Criteria}} & \multicolumn{1}{c}{\textbf{Agreement}} & \multicolumn{1}{c}{\textbf{Precision}} & \multicolumn{1}{c}{\textbf{Recall}} & \multicolumn{1}{c}{\textbf{F1}} \\
\hline \multicolumn{5}{l}{\textit{Existing Baseline. (* from original paper)}}  \\ 
\multicolumn{5}{l}{\textit{Base Models.}} \\
\quad Qwen2.5-3B-Instruct  & 74.27 & 74.47 & 73.89 & 73.96 \\
\quad Qwen2.5-7B-Instruct  & 74.50 & 74.49 & 74.57 & 74.48 \\
\quad Deepseek-R1 & 83.48 & 83.53 & 83.31 & 83.38 \\
\hdashline
\multicolumn{5}{l}{\textit{Ours.}} \\
\quad JudgeLRM-3B & 82.66 & 82.97 & 82.94 & 82.66 \\
\quad JudgeLRM-7B  & \textbf{83.56} & \textbf{83.58} & \textbf{83.41} & \textbf{83.47}\\
\quad\quad - w/o. judge\_score\_reward& 82.44 & 82.88 & 82.77 & 82.44 \\
\bottomrule
\end{tabular}
}
\end{center}
\captionsetup{type=table}
\vspace{-0.5em}
\caption{Performance on PandaLM testset excluding ties.}
\label{tab:excludetie}
\end{table}

\section*{Case Study 1}

Fig. \ref{fig:criteria_evaluation_part1} show a case of question, and Fig. \ref{fig:criteria_evaluation_our3b},\ref{fig:criteria_evaluation_our3b_2} shows the result from JudgeLRM-3B. Fig. \ref{fig:criteria_evaluation_2_deepseekr1} shows
the result of Deepseek-R1. Fig. \ref{fig:criteria_evaluation_gold} shows the gold answer provided by JudgeLM from GPT-4.

\begin{figure*}[!h]
\centering
\begin{tcolorbox}[colback=blue!5!white, colframe=blue!70!black, title=Case Question 1, width=1.0\textwidth, sharp corners=south]
\texttt{[Question]}\\I'm trying to find out how to get my ex-boyfriend back. Can you help me?\\\\\texttt{[Assistant 1's Answer]}\\I'm sorry to hear that you're going through a tough time. However, as an AI language model, I cannot provide relationship advice. You may want to seek guidance from a trusted friend or a professional counselor who can offer you tailored advice and support based on your unique situation. I wish you the best of luck.\\\\\texttt{[Assistant 2's Answer]}\\I'm sorry to be the one to tell you this but, no, we can't help you get your ex-boyfriend back.\\Our site is a free resource that offers advice and tips on how to meet new people and make new friends.\\If you're looking for advice on how to get an ex-boyfriend back, we cannot help you.\\Your ex-boyfriend is still with you, and we would therefore advise you to use this time to try to find out what went wrong in your relationship. You have a good chance of getting him back, but we cannot advise you on how to do this.\\We're here to help you find friendship and romance, not to help you get back with your ex-boyfriend.\\We wish you all the best with this and hope we've helped you find a friend here!The Best-Selling and Award-Winning Author of THE EAGLE'S GIFT Trilogy and The Secret Circle.\\Do you want to know if he still loves you?\\Get the answer here for free.\\Are you ready to meet other singles? Get started now, it's free!\\What to Do Now that He Ignored You?\\

\end{tcolorbox}

\caption{Case Question 1.}
\label{fig:criteria_evaluation_part1} 
\end{figure*}

\begin{figure*}[!h]
\centering
\begin{tcolorbox}[colback=blue!5!white, colframe=blue!70!black, title=An Example of Answer from JudgeLRM-3B, width=1.0\textwidth, sharp corners=south]
\texttt{<think>}

For Assistant 1's response, the tone and content are appropriate for providing guidance and support in a respectful manner, while emphasizing limitations of the AI service. It maintains a courteous and informative approach by advising the user to seek help from friends or professionals rather than attempting to directly influence their ex-partner's actions. 

The criteria considered include helpfulness - it gives helpful advice about seeking support; relevance - it stays focused on the query and suggests appropriate resources; accuracy - it avoids any misleading information; and level of detail - it provides sufficient information without being overly verbose.

In terms of length, the response remains concise and straightforward. It avoids unnecessary tangents or irrelevant information, making the message clear and direct. It also adheres to the boundaries set by the capabilities of an AI language model. 

Regarding Assistant 2's response, it contains several issues including repetition, redundancy, inappropriate suggestions, and the inclusion of promotional elements which are unrelated to the core inquiry. These aspects make the answer less helpful, relevant, accurate, and detailed compared to Assistant 1’s response. 

Assistant 2 employs a confrontational tone, which is inconsistent with the initial polite greeting and seems out of place given the context of seeking advice. Additionally, the answer does not adhere strictly to the criteria mentioned: it fails to provide substantial, direct advice and instead veers into irrelevant topics such as promoting services and speculating about past relationships. Furthermore, the inclusion of personal opinions, endorsements, and authorship details detracts from the informational value and distracts from the primary objective of providing advice. Lastly, the inclusion of unrelated statements like "Are you ready to meet other singles?" and "What to Do Now that He Ignored You?" does not add to the quality of the advice and appear more suited for a dating app rather than a relationship recovery guide. \\\\\texttt{</think>}\\\\\\\texttt{<answer>}8\texttt{</answer>}\texttt{<answer>}4\texttt{</answer>}
\end{tcolorbox}
\caption{An example of thinking and answer from Judgelrm 3B after training 500 steps.}
\label{fig:criteria_evaluation_our3b} 
\end{figure*}

\begin{figure*}[!h]
\centering
\begin{tcolorbox}[colback=blue!5!white, colframe=blue!70!black, title=An Example of Answer from JudgeLRM-3B, width=1.0\textwidth, sharp corners=south]
\texttt{<think>}\\Okay, so I need to evaluate the performance of Assistant 1 and Assistant 2 based on the user's question about getting an ex-boyfriend back. The criteria are helpfulness, relevance, accuracy, and level of detail. Let me break it down step by step.\\\\First, looking at Assistant 1's response. They start by expressing sympathy, which is good for empathy. Then they state they can't provide relationship advice and suggest seeking help from friends or professionals. That's responsible because giving relationship advice can be tricky and possibly harmful. However, they don't offer any resources or steps, just a referral. So, helpfulness is limited but honest. Relevance is okay because they address the user's intent but don't engage further. Accuracy is good because they're correct in directing to real-world help. Level of detail is low since there's no actionable advice.\\\\Now, Assistant 2. Their answer starts by saying they can't help, which is direct. They mention their site is for meeting new people, which is a bit off-topic since the user wants to reconcile. Then they suggest reflecting on the relationship's issues, which is somewhat helpful. But then the response gets messy with promotional content about books, a quiz, and dating site plugs. The irrelevant links and self-promotion lower the score. There's some attempt to advise but it's buried in distractions. Accuracy is questionable because parts like "Your ex-boyfriend is still with you" might be incorrect. Relevance suffers due to the unrelated links. Detail is inconsistent\u2014some advice but mostly filler.\\\\Comparing both: Assistant 1 is concise, empathetic, and correctly refers to professionals but lacks details. Assistant 2 tries to give advice but includes irrelevant content and promotions, making it less helpful and accurate. Assistant 2's promotional aspects and possible inaccuracies drag down its score. Assistant 1 is more focused and responsible, though not detailed. So scores would be Assistant 1 around 6 for being relevant and accurate but not detailed. Assistant 2 lower, maybe 3, due to irrelevant content and lower accuracy.\\\\\texttt{</think>}\\\\\\\texttt{<answer>}7\texttt{</answer>}\texttt{<answer>}3\texttt{</answer>}
\end{tcolorbox}
\caption{Case 1 example of thinking and answer from Judgelrm 3B.}
\label{fig:criteria_evaluation_our3b_2} 
\end{figure*}

\begin{figure*}[!h]
\centering
\begin{tcolorbox}[
  colback=blue!5!white,
  colframe=blue!70!black,
  title=The reasoning of Deepseek R1,
  width=\textwidth,
  sharp corners=south,
  fontupper=\small,
]
\texttt{Example 1 Answer:}\\[1ex]
\texttt{<think>}\\
Okay, I need to evaluate the two assistants' answers based on the criteria: helpfulness, relevance, accuracy, and level of detail. Let me break it down step by step. First, the user's question is about getting their ex-boyfriend back. They're seeking advice, so the assistants' responses should address that need appropriately.

Starting with Assistant 1: The response is empathetic, acknowledging the user's feelings. They state they can't provide relationship advice and suggest seeking help from friends or professionals. That's responsible, as giving relationship advice can be tricky and potentially harmful. However, while it's honest and cautious, it might not be very helpful because it doesn't offer any actionable steps. Relevance is there since they address the request, but accuracy is neutral because they don't provide incorrect info. Level of detail is low because there's no specific guidance, just a referral elsewhere.

Now Assistant 2: Their answer starts by directly refusing help but then gets confusing. They mention their site is for meeting new people, which isn't what the user asked. There's a contradiction: first saying they can't help, then suggesting the user has a good chance but can't advise. This inconsistency might be misleading. Also, the inclusion of promotional links (free answers, meeting singles) is irrelevant and possibly distracting. The advice to "find out what went wrong" is somewhat relevant but buried in a lot of off-topic content. The accuracy is questionable here because parts of the response are contradictory and include unsolicited ads. The level of detail is mixed; some parts try to address the issue but get sidetracked.

Comparing both, Assistant 1 is more straightforward and honest, albeit brief. Assistant 2 tries to give some advice but muddies it with irrelevant content and self-promotion, which could be unhelpful or even harmful. However, Assistant 2 does touch on evaluating the relationship's issues, which is slightly more detailed but not effectively communicated.

In terms of helpfulness, Assistant 1 gets points for directing to proper help, while Assistant 2's advice is mixed with spam. Relevance: Assistant 1 stays on topic better. Accuracy: Assistant 1 is more accurate by not giving potentially wrong advice. Detail: Neither is highly detailed, but Assistant 2's extra content is not useful. So, scores would be Assistant 1 around 6-7, Assistant 2 lower, maybe 3-4.

\texttt{<answer>}7\texttt{</answer>}\texttt{<answer>}3\texttt{</answer>}\\[1ex]
\\ 
\texttt{Example 2 Answer:}\\[1ex]
\texttt{<think>}\\
Okay, let's start by looking at the user's question. They want to get their ex-boyfriend back and are asking for help. Now, I need to evaluate both Assistant 1 and Assistant 2's responses based on the criteria: helpfulness, relevance, accuracy, and level of detail. Starting with Assistant 1: The response is polite and empathetic. They mention not being able to provide relationship advice and suggest seeking help from friends or professionals. This is responsible, but maybe a bit too brief. It doesn't address the user's specific request directly beyond declining to help. However, it's accurate in stating their limitations and points the user to other resources. The relevance is there because they acknowledge the user's struggle, but the helpfulness is limited since they don't offer any actionable advice. Level of detail is low—just a short paragraph. Now Assistant 2: The response starts by saying they can't help get the ex back but then contradicts that by saying "we cannot advise you on how to do this." Wait, that's confusing. They mention their site is for meeting new people, which is not what the user asked. Then there's a lot of promotional content like ads for books, quizzes, and dating sites. The answer includes irrelevant links and self-promotion, which isn't helpful. The accuracy is questionable because parts of the response are conflicting (saying they can't help but also hinting there's a chance). The relevance is low because they divert the user to other services instead of addressing the request. Also, the level of detail is poor as they don't provide any real advice but add unnecessary elements.

Comparing both, Assistant 1 is honest, stays within their role, and directs to appropriate help, though it's not detailed. Assistant 2 is more problematic—unclear, includes ads, and isn't focused on the user's needs. Assistant 1 might get a higher score for being relevant and accurate, even if brief. Assistant 2's score would be lower due to irrelevant content and inaccuracies.

\texttt{</think>}\\[2ex]
\texttt{<answer>}6\texttt{</answer>}\texttt{<answer>}3\texttt{</answer>}
\end{tcolorbox}
\caption{The reasoning of Deepseek-R1 of Case 1.}
\label{fig:criteria_evaluation_2_deepseekr1}
\end{figure*}

\begin{figure*}[!t]
\centering
\begin{tcolorbox}[colback=blue!5!white, colframe=blue!70!black, title=Gold Response of Case 1 from GPT-4, width=1.0\textwidth, sharp corners=south]
    8 4\\Assistant 1 provided a compassionate and professional response, acknowledging the user's emotional state and suggesting seeking advice from trusted friends or professional counselors. Although it didn't provide a direct answer to the user's question, it was still helpful and relevant in the context of the question. On the other hand, Assistant 2's response was confusing and inconsistent. It started by stating that it couldn't help the user get her ex-boyfriend back, then suggested that the user's ex-boyfriend was still with her, which contradicts the user's question. It also included irrelevant promotional content about meeting other singles and a best-selling author, which was not related to the user's question. Therefore, Assistant 1's response was more helpful, relevant, and accurate than Assistant 2's.
\end{tcolorbox}
\caption{Gold Response of Case 1 from Gpt-4.}
\label{fig:criteria_evaluation_gold} 
\end{figure*}

\section*{Case study 2}
Fig. \ref{fig:r1_judge_case2} show another case of question, and Fig. \ref{fig:case2our7b} shows the result from JudgeLRM-7B. Fig. \ref{fig:criteria_evaluation_r1_2} shows
the result of Deepseek-R1. Fig. \ref{fig:criteria_evaluation_2_gpt4} shows the gold answer provided by JudgeLM from GPT-4.  
\begin{figure*}[!h]
\centering
\begin{tcolorbox}[colback=blue!5!white, colframe=blue!70!black, title=Case Question 2, width=1.0\textwidth, sharp corners=south]
\texttt{[Question]}\\I've heard that some people can have high blood pressure without having any symptoms. Is this true?\\\\\texttt{[Assistant 1's Answer]}\\Healthcare provider: Yes, this is true. High blood pressure is a condition in which the blood pressure in your arteries is too high. It can be caused by a variety of factors, such as stress, diet, and exercise.\\\\Patient: I'm aware of the factors that can increase blood pressure, but why do some people have high blood pressure without any symptoms?\\\\Healthcare provider: It's possible that some people with high blood pressure may not have any symptoms, especially if the blood pressure is not very high. However, high blood pressure puts you at risk of developing heart disease and stroke, so it's important to have your blood pressure checked regularly and to make any necessary lifestyle changes to keep it under control.\\\\Patient: I understand. How can I lower my blood pressure?\\\\Healthcare provider: There are several things you can do to lower your blood pressure. One of the most important things is to make lifestyle changes, such as:\\\\*   Eating a healthy diet that is rich in fruits, vegetables, and whole grains\\*   Exercising regularly\\*   Reducing stress\\\\\texttt{[Assistant 2's Answer]}\\And if so, how can I lower my risk of this disease? Thank you!

\end{tcolorbox}
\caption{Case Question 2.}
\label{fig:r1_judge_case2} 
\end{figure*}

\section*{Figure of Response Length}
\begin{figure}[h!] 
\centering
\begin{minipage}{0.48\linewidth}
    \centering
    \includegraphics[width=\linewidth]{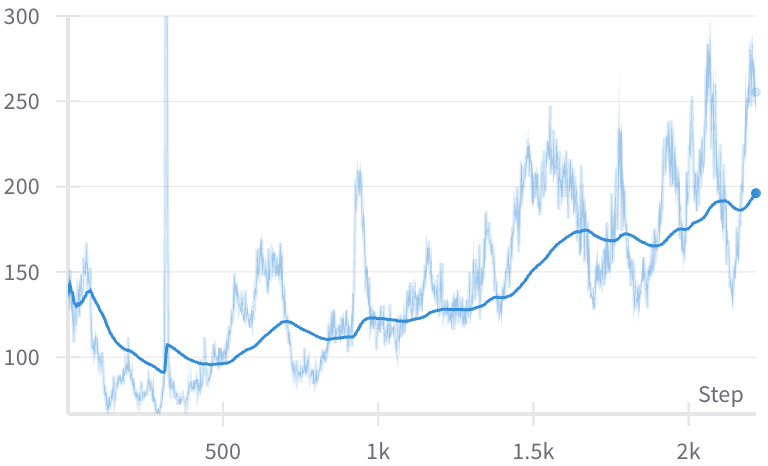} 
    \label{fig:aime}
\end{minipage}
\hspace{0.02\linewidth}
\begin{minipage}{0.48\linewidth}
    \centering
    \includegraphics[width=\linewidth]{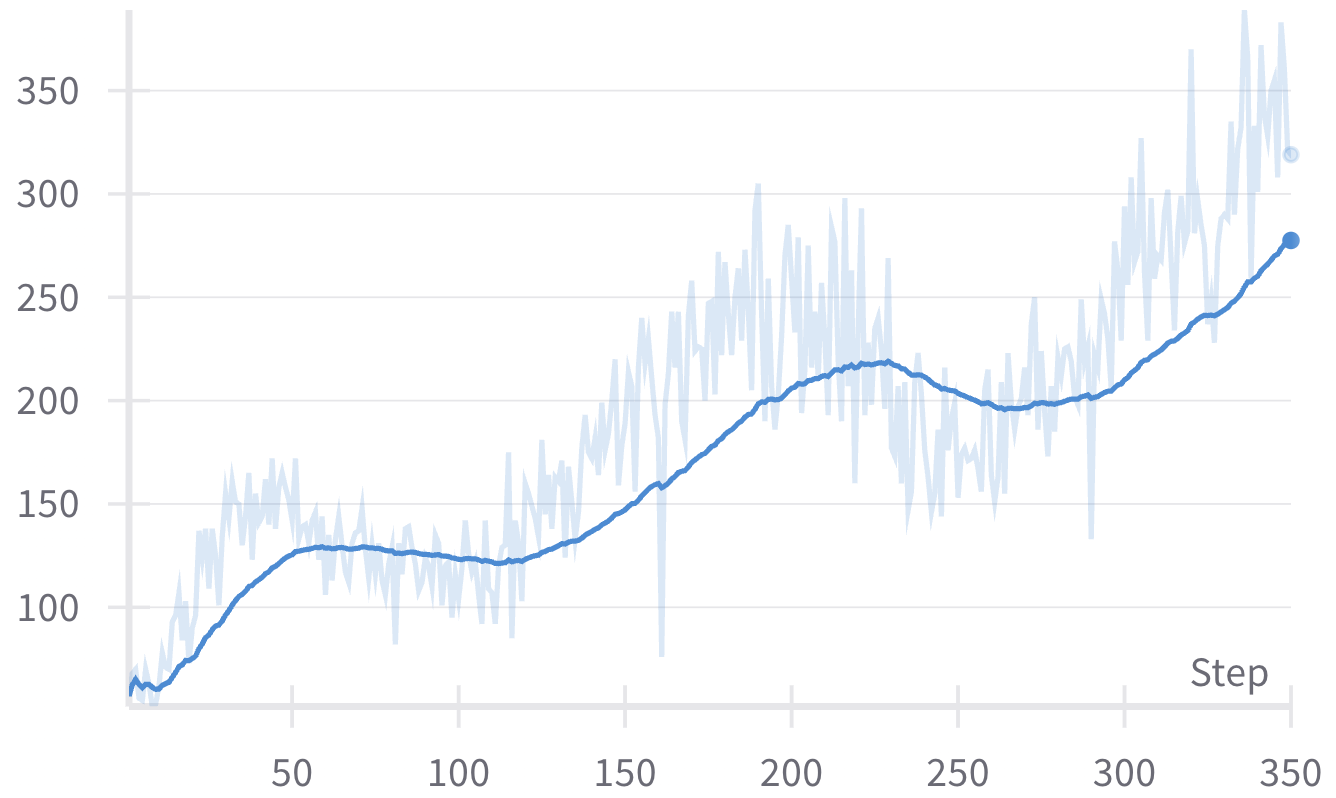} 
    \label{fig:amc}
\end{minipage}
\vspace{-4mm}
\caption{Response Length of JudgeLRM-3B and JudgeLRM-7B by Steps.}
\label{fig:step}
\end{figure}

Fig. \ref{fig:step} shows the response length of JudgeLRM-3B and JudgeLRM-7B by steps.

\begin{figure*}[h!]
\centering
\begin{tcolorbox}[
  colback=blue!5!white, 
  colframe=blue!70!black, 
  title=The reasoning of JudgeLRM-7B, 
  width=\textwidth, 
  sharp corners=south,
  fontupper=\small, 
  boxrule=1pt
]
\texttt{<think>} I will evaluate Assistant 1 and Assistant 2 based on the criteria of helpfulness, relevance, accuracy, and level of detail.\\[1ex]

Assistant 1 starts with a statement about their own paleness, which is a bit whimsical and not necessary in the context. They then provide information about avoiding sunburn, which doesn't seem to answer the asked question—about tanning. They mention alternatives like a tanning bed, sunscreen, and protective clothing as ways to avoid burning. This information is generally useful and relevant. However, their response is primarily focused on preventing sunburn rather than providing information on how to tan without burning. The explanation can be more practical and customer-oriented, providing a clearer understanding to someone looking for tips on tanning.\\[1ex]

Assistant 2 explains that there are multiple ways to get a tan while maintaining skin health. They suggest using sunscreen, which, while protective, can indeed help with tanning. They also advise to use a tanning bed carefully, which seems a fair inclusion in a variety of methods. Both assistants mention protective clothing, which is probably the most effective method to avoid sunburn and to ensure safe tanning. They both also give a brief mention of times to avoid direct sunlight for best results.\\[1ex]

Assistant 2's response, however, provides a more comprehensive and clear explanation of different methods to get a tan without burning, while also providing context on how to protect your skin from UV rays. In addition, they explore the relation between tanning and vitamin D production, pointing out that getting a tan might be healthy for a general audience. Their response seems to be more to the point and fits the asked question a little better, also giving more contextual information that could be valuable for the reader.\\[1ex]

Although both Assistants seem to be accurate, in the sense that the advice given does not convey any apparent misinformation, Assistant 2 provides more detailed information about individual ways to get a tan while also explaining the potential benefits and health aspects, thus more relevant to a five-year-old than Assistant 1.\\[1ex]

In terms of helpfulness, both provide useful suggestions. Assistant 1 is a tad less clear and focused, while Assistant 2’s points are slightly easier to understand and can be clearly applied to the reader's context. Assistant 2 also provides educational information about skin health and tanning, found in their reference to vitamin D benefits, which can be slightly helpful.\\[1ex]

Assistant 1’s response is more conversational, and although it provides general information, the tone seems more casual and could be confusing, while Assistant 2's is more structured and detailed, providing a clear and helpful mixture of information and advice.\\[1ex]

\texttt{</think>}\\[2ex]
\texttt{<answer>}6\texttt{</answer>}\texttt{<answer>}9\texttt{</answer>}
\end{tcolorbox}
\caption{The reasoning of JudgeLRM-7B in Case 2.}
\label{fig:case2our7b} 
\end{figure*}

\begin{figure*}[!h]
\centering
\begin{tcolorbox}[
  colback=blue!5!white,
  colframe=blue!70!black,
  title=Deepseek-R1’s Thinking and Answer of Case 2, 
  width=\textwidth,
  sharp corners,      
  fontupper=\footnotesize, 
]
Okay, so the user is asking if it's true that people can have high blood pressure without symptoms. Let's look at both assistants' answers to evaluate their performance.\\\\Starting with Assistant 1's answer. The healthcare provider responds affirmatively, which is correct. They mention that high blood pressure can be asymptomatic, especially if not very high. That's accurate. They also note the risks like heart disease and stroke, which is good. The patient then asks how to lower blood pressure, and the provider lists diet, exercise, and stress reduction. But wait, the answer seems cut off. The formatting is a bit odd with the patient and provider dialogue. The healthcare provider's advice is correct but maybe lacks some detail. For example, they could mention reducing sodium intake, limiting alcohol, or medications if needed. Also, the response to the second patient question about lowering blood pressure is a bit brief. The answer is relevant and accurate but could be more detailed.\\\\Now, Assistant 2's answer. Wait, the user's question ends with "And if so, how can I lower my risk of this disease? Thank you!" but Assistant 2's answer seems incomplete. Actually, looking back, the user's message might be a continuation. Wait, the user's message includes two parts: the initial question and then a follow-up. However, in the provided dialogue, Assistant 2's answer is just "And if so, how can I lower my risk of this disease? Thank you!" which seems like a repetition of the user's follow-up question, not an actual answer. That might be a formatting error. Wait, maybe the user's input was split, and Assistant 2's answer is actually the part after the user's follow-up. But looking at the structure, after the user's first question, both assistants' answers are presented. Wait, the user's message includes the question and then two assistants' answers. The first assistant's answer is a dialogue between healthcare provider and patient, addressing the initial question and follow-up. The second assistant's answer seems to be only "And if so, how can I lower my risk of this disease? Thank you!" which is not an answer but a repetition of the user's follow-up. That must be a mistake in the input formatting. Alternatively, maybe Assistant 2's answer is missing. Wait, the way it's structured: after the user's question, the user's message includes the question and then the two assistants' answers. So Assistant 1's answer is the dialogue, and Assistant 2's answer is the line starting with "And if so..." which is actually part of the user's message. Wait, no, looking at the exact structure:\\\\The user's message is:\\\\\texttt{[Question]}\\I've heard that some people can have high blood pressure without having any symptoms. Is this true?\\\texttt{[Assistant 1's Answer]}\\... dialogue ...\\\texttt{[Assistant 2's Answer]}\\And if so, how can I lower my risk of this disease? Thank you!\\\\Ah, so Assistant 2's answer is just that line, which is actually the user's follow-up question, not an answer. That's a problem. So Assistant 2's answer is incomplete or incorrectly formatted. It seems like Assistant 2's answer is just repeating the user's own follow-up question, which doesn't make sense. Therefore, Assistant 2's answer is not providing any useful information. It's either a mistake or a placeholder. So in terms of scoring, Assistant 1 provided a relevant and accurate answer, albeit a bit brief and with some formatting issues (like the dialogue structure which might be confusing). Assistant 2's answer is nonsensical as it's just repeating the user's question. Therefore, Assistant 2 gets a low score.\\\\But wait, maybe I'm misinterpreting. Let me check again. The user's message includes the initial question and then two assistant answers. The first answer is the dialogue, and the second is that line. So Assistant 2's answer is just "And if so..." which is not an answer. So Assistant 2 failed to provide a proper response. Therefore, Assistant 1 gets points for correctness and relevance but loses some for lack of detail. Assistant 2's answer is not an answer, so it's irrelevant and unhelpful. So scores: Assistant 1 maybe a 7 (correct but could be more detailed), Assistant 2 a 1 or 2 for not answering.\\\texttt{</think>}\\\texttt{<answer>}7\texttt{</answer>}\texttt{<answer>}2\texttt{</answer>}"
\end{tcolorbox}
\caption{Example of Deepseek-R1's Thinking and Answer of Case 2.}
\label{fig:criteria_evaluation_r1_2} 
\end{figure*}

\begin{figure*}[!h]
\centering
\begin{tcolorbox}[colback=blue!5!white, colframe=blue!70!black, title=Gold Response of Case 2 from GPT-4, width=1.0\textwidth, sharp corners=south]
9 1\\Assistant 1 provided a comprehensive, accurate, and relevant answer to the user's question. The assistant explained what high blood pressure is, why some people might not have symptoms, and the risks associated with it. The assistant also provided detailed advice on how to lower blood pressure, which was very helpful. Therefore, Assistant 1 receives a score of 9.\\\\On the other hand, Assistant 2's response was not relevant to the user's question. It seems like the assistant misunderstood the question and provided an incomplete response. The assistant did not provide any information about high blood pressure or how to lower it. Therefore, Assistant 2 receives a score of 1.
\end{tcolorbox}
\caption{Gold response of Case 2 from GPT-4 (Provided in PandaLM dataset) (No reasoning from gpt-4).}
\label{fig:criteria_evaluation_2_gpt4} 
\end{figure*}

\end{document}

%% file: math_commands.tex
\usepackage{amsmath,amsfonts,bm}

\def\eqref#1{equation~\ref{#1}}

\def\1{\bm{1}}

\DeclareMathAlphabet{\mathsfit}{\encodingdefault}{\sfdefault}{m}{sl}
\SetMathAlphabet{\mathsfit}{bold}{\encodingdefault}{\sfdefault}{bx}{n}

